%% file: tmi.tex
\documentclass[journal]{IEEEtran}
\usepackage{color}
\usepackage[dvipsnames]{xcolor}
\usepackage{xcolor}
\usepackage{booktabs}

\usepackage{tmi}

\usepackage{amsmath,amssymb,amsfonts}
\usepackage{algorithmic}
\usepackage{graphicx}
\usepackage{xspace}
\usepackage{subcaption}
\usepackage{bbold}
\usepackage{textcomp}
\usepackage{mathtools}

\usepackage{multirow}

\usepackage[numbers, compress]{natbib}
\input{math_commands}

\usepackage{array}

\def\BibTeX{{\rm B\kern-.05em{\sc i\kern-.025em b}\kern-.08em
    T\kern-.1667em\lower.7ex\hbox{E}\kern-.125emX}}
\makeatletter
\DeclareRobustCommand\onedot{\futurelet\@let@token\@onedot}
\def\@onedot{\ifx\@let@token.\else.\null\fi\xspace}
\def\eg{\emph{e.g}\onedot} 

\def\ie{\emph{i.e}\onedot}

\def\vs{\emph{vs}\onedot}

\def\etal{\emph{et al}\onedot}
\makeatother

\colorlet{CLRBlue}{black}
\colorlet{CLRBlue2}{black}

\begin{document}

\title{Cyclical Self-Supervision  for Semi-Supervised Ejection Fraction Prediction from Echocardiogram Videos}
\author{Weihang Dai, Xiaomeng Li, \IEEEmembership{Member, IEEE}, Xinpeng Ding, Kwang-Ting Cheng, \IEEEmembership{Fellow, IEEE}

\thanks{Manuscript received XX XX, 2022. 

This work was supported by a grant 
from the National Natural Science Foundation of China / Research Grants Council Joint Research Scheme under Grant \#N\_HKUST627, 
a grant from the Shenzhen Municipal Central Government Guides Local Science
and Technology Development Special Funded Projects under Grant
2021Szvup139 
and a grant from HKUST-BICI Exploratory Fund under HCIC-004.

W. Dai is with the Department of Computer Science and Engineering, The Hong Kong University of Science and Technology, Hong Kong SAR, China (e-mail:wdai03@gmail.com).

X. Li is with the Department of Electronic and Computer Engineering, The Hong Kong University of Science and Technology, Hong Kong SAR, China, and also with The Hong Kong University of Science and Technology Shenzhen Research Institute (e-mails: eexmli@ust.hk)
(Corresponding author: Xiaomeng Li.)

X. Ding is with the Department of Electronic and Computer
Engineering, The Hong Kong University of Science and Technology, Hong Kong SAR, China (e-mail:xdingaf@connect.ust.hk). 

K.T. Cheng is with the Department of Computer Science and Engineering and the Department of Electronic and Computer Engineering, The Hong Kong University of Science and Technology, Hong Kong SAR, China (e-mail: timcheng@ust.hk). 
} 
\thanks{Copyright (c) 2022 IEEE. Personal use of this material is permitted. Permission from IEEE must be obtained for all other uses, including reprinting/republishing this material for advertising or promotional purposes, collecting new collected works for resale or redistribution to servers or lists, or reuse of any copyrighted component of this work in other works.}
}

\maketitle

\newcommand{\xmli}[1]{{\color{blue}{[XM: #1]}}}
\newcommand{\wdai}[1]{{\color{red}{[WD: #1]}}}
\newcommand{\wdairv}[1]{{\color{CLRBlue}{#1}}}
\newcommand{\wdairvb}[1]{{\color{CLRBlue2}{#1}}}

\begin{abstract}

Left-ventricular ejection fraction (LVEF) is an important indicator of heart failure.
Existing methods for LVEF estimation from video require large amounts of annotated data to achieve high performance, \eg using 10,030 labeled echocardiogram videos to achieve \wdairv{mean absolute error (MAE)} of 4.10. Labeling these videos is time-consuming however and limits potential downstream applications to other heart diseases.
This paper presents the first semi-supervised approach for LVEF prediction.
Unlike general video prediction tasks, LVEF prediction is specifically related to changes in the left ventricle (LV) in echocardiogram videos. By incorporating knowledge learned from predicting LV segmentations into LVEF regression, we can provide additional context to the model for better predictions. 
To this end, we propose a novel Cyclical Self-Supervision (CSS) method for learning video-based LV segmentation, which is motivated by the observation that the heartbeat is a cyclical process with temporal repetition. 
Prediction masks from our segmentation model can then be used as additional input for LVEF regression to provide spatial context for the LV region.
We also introduce teacher-student distillation to distill the information from LV segmentation masks into an end-to-end LVEF regression model 
that only requires video inputs. 
Results show our method outperforms alternative semi-supervised methods and can achieve MAE of 4.17, which is competitive with state-of-the-art supervised performance, using half the number of labels.
Validation on an external dataset also shows improved generalization ability from using our method.

\end{abstract}


\section{Introduction}
\label{sec:introduction}

\begin{figure}%
\centering
\begin{subfigure}{\columnwidth}
\centering
\includegraphics[width=0.9 \columnwidth]{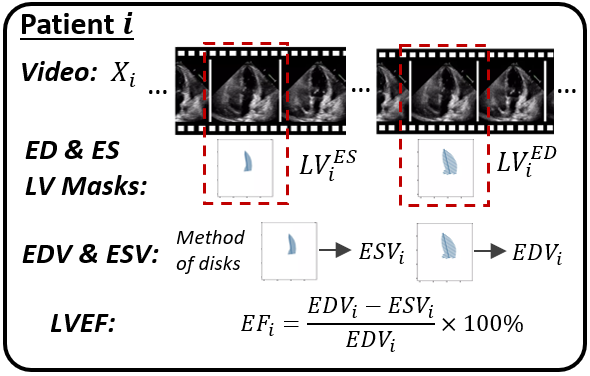}%
\caption{Example of data collection process for LVEF}%
\label{setting_a}%
\end{subfigure}\hfill%
\begin{subfigure}{\columnwidth}
\centering
\includegraphics[width=0.85 \columnwidth]{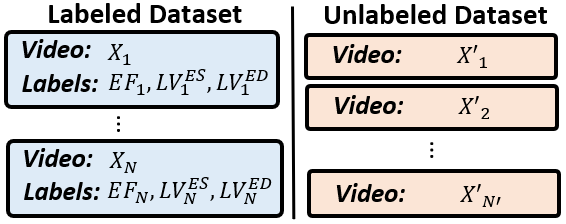}%
\caption{Problem setting for semi-supervised LVEF prediction}%
\label{setting_b}%
\end{subfigure}\hfill%

\caption{ \textbf{(a)} The labeling process for \wdairv{left-ventricular ejection fraction} (LVEF) involves first identifying representative \wdairv{end-diastole} (ED) and \wdairv{end-systole} (ES) frames from the echocardiogram video sequence. The \wdairv{left ventricle} (LV) is then segmented from these two frames to predict \wdairv{end-diastole volume} (EDV) and \wdairv{end-systole volume} (ESV) using method of disks~\cite{bamira2018imaging}. LVEF is calculated as their percentage difference. \textbf{(b)} For semi-supervised LVEF prediction, we aim to use labeled sequences together with completely unlabeled video sequences for model training to reduce annotation requirements. 
}
\label{setting}
\end{figure}

\begin{figure*}
\centering

\begin{subfigure}{\columnwidth}
\centering
\includegraphics[width=0.99 \columnwidth]{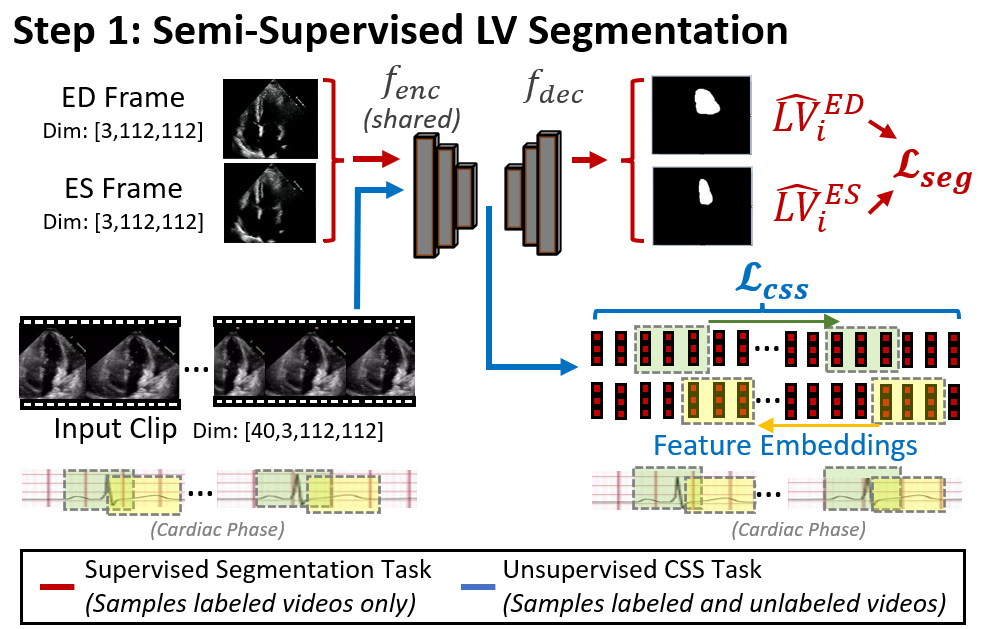}%
\caption{}%
\label{flow_a}%
\end{subfigure}\hfill%
\begin{subfigure}{\columnwidth}
\centering
\includegraphics[width=0.99 \columnwidth]{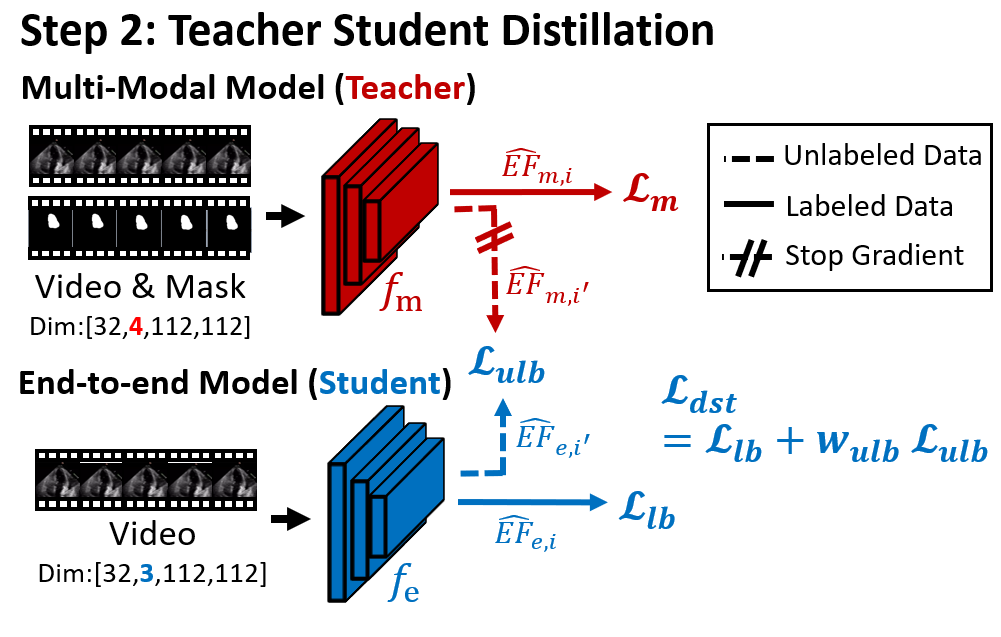}%
\caption{}%
\label{flow_b}%
\end{subfigure}\hfill%

\caption{We approach semi-supervised LVEF prediction in two steps. 
\textbf{(a) Step 1}: We perform semi-supervised LV segmentation to obtain LV mask predictions for multi-input LVEF regression. \wdairv{We sample a batch of ED and ES frames (\textcolor{red}{red}), which have dimensions $3\times112\times112$, to calculate segmentation loss \wdairv{($\mathcal{L}_{seg}$)} for supervised segmentation. We \textit{separately} sample an input clip of 40 frames (\textcolor{blue}{blue}), which covers a minimum of two cardiac cycles, for enforcing cyclical consistency through minimizing Cyclical Self-Supervision loss ($\mathcal{L}_{css}$). This is based on the observation that the heartbeat is a cyclically repeating process, and features should be cyclically consistent. 
The two tasks share the same model encoder and are jointly trained.} 
\textbf{(b) Step 2:} We train a multi-input model $f_m$ (\textcolor{red}{red}) \wdairv{using mean squared error (MSE) loss ($\mathcal{L}_m$)} on video concatenated with segmentation predictions, which has 4 input channels. The spatial context from LV mask predictions is then distilled into an end-to-end model for 3 channel video inputs, $f_e$ (\textcolor{blue}{blue}), by using $f_m$ as a teacher model in teacher-student distillation. \wdairv{Distillation loss ($\mathcal{L}_{dst}$) is calculated based on MSE of labeled ($\mathcal{L}_{lb}$) and unlabeled samples ($\mathcal{L}_{ulb}$) with unlabeled loss weighting of $w_{ulb}$.  }
}
\label{flow}
\end{figure*}

Left ventricular ejection fraction (LVEF) is one of the most commonly reported measures of left ventricular (LV) systolic function and is used to diagnose cardiac disease~\cite{maeder2009heart,abdi2017automatic,hughes2021deep}. It is calculated as the percentage volume change between the end-diastole (ED) and end-systole (ES) phases of the heartbeat, which corresponds to maximum and minimum volumes of the LV respectively. Values outside the normal range of 40\% to 70\% indicate potential heart failure or cardiomyopathy~\cite{maeder2009heart}. Manual measurement of LVEF involves identifying representative ED and ES frames from the echocardiogram sequence, tracing the LV chamber to estimate volume, and calculating the percentage difference~\cite{bamira2018imaging,ouyang2019echonet}; see Fig.~\ref{setting_a}. This process is time-consuming and suffers from large inter-observer variation, thus motivating the development of automated solutions~\cite{ouyang2020video,cole2015defining}.

One approach for LVEF prediction is to estimate end-diastole volume (EDV) and end-systole volume (ESV) separately based on predicted LV segmentations~\cite{jafari2021deep,reynaud2021ultrasound,wei2020temporal}. These methods require accurate identification of ED and ES frames however since errors from frame selection and volume estimation can propagate to the final LVEF estimate~\cite{wei2020temporal,yuan2021systematic}. Current state-of-the-art methods achieve decent results by directly regressing LVEF from video inputs, an end-to-end approach. However, video understanding requires large amounts of labeled data. For example, Ouyang~\etal~\cite{ouyang2020video} and Dai~\etal~\cite{dai2021adaptive} develop LVEF regression models using 10,030 annotated echocardiogram videos, which are costly to label. Moreover, the requirement for extensive annotations makes it challenging to generalize the methods for diagnosing related diseases, such as elevated \wdairv{B-type natriuretic peptide} (BNP) levels or LV hypertrophy~\cite{hughes2021deep,ghorbani2020deep}.
To this end, we propose a semi-supervised method to predict LVEF from echocardiogram videos using labeled and unlabeled data. Fig.~\ref{setting_b} shows our problem setting.

To our best knowledge, this is the first paper exploring semi-supervised LVEF prediction from echocardiogram videos.
We observe that, unlike natural video understanding, LVEF is directly related to changes in LV volume.
We can therefore provide additional context for LVEF video regression by concatenating LV segmentation masks with video frames, as this helps highlight the relevant region. 
Based on this insight, we approach semi-supervised LVEF prediction in two steps: (1) we train a semi-supervised video segmentation model to generate LV segmentation predictions for individual frames; (2) we then distill the information provided by segmentation masks into an end-to-end LVEF regression model that directly performs inference using raw video inputs only. 

For semi-supervised video-based LV segmentation, we introduce a novel cyclical self-supervision (CSS) method that enforces feature similarity based on the cyclical nature of the heartbeat. This is based on the observation that the heart is \textit{visually similar during the same phases of a cycle}, such as at the start of contraction, and therefore \textit{should have similar features}. 
Experiments show that segmentation masks generated using CSS improve LVEF video regression when concatenated with video as additional input and outperform alternative segmentation methods (see Table~\ref{tab:ablat_seg}). 
For model distillation, we propose teacher-student distillation to distill information captured by segmentation predictions into an end-to-end model. Our method uses pseudo-labels on unlabeled videos for distillation, which also allows additional data to be used in training and further improves predictions (see Table~\ref{tab:lab_available}). 
An illustration of our workflow is shown in Fig~\ref{flow}.  

Our method outperforms existing semi-supervised approaches for video learning, achieving 4.90 mean absolute error (MAE) with one-eighth of available labels (see Table~\ref{tab:sota}). We also achieve 4.17 MAE using half the available labels, which is competitive with 4.10 MAE achieved using the supervised approach by Ouyang \etal~\cite{ouyang2020video} on a fully labeled dataset. External validation further shows that our semi-supervised method has better generalization ability and outperforms state-of-the-art fully supervised models on data from a different hospital (see Table~\ref{tab:external_val}). Our key contributions are:

\begin{itemize}
    \item 
    This paper is the first work to propose an annotation-efficient approach to LVEF prediction from echocardiogram videos using labeled and unlabeled video sequences.
    
    \item 
    Given the cyclical nature of echocardiograms, we introduce a novel cyclical self-supervision (CSS) method for semi-supervised video segmentation that enforces feature similarity based on temporal cyclicality.
    
    \item 
    Because LVEF is determined by changes in the volume of the LV, we introduce a novel teacher-student distillation method to distill spatial context from LV segmentation masks into an end-to-end LVEF video regression model.
    
    \item 
    Our method outperforms existing semi-supervised video learning techniques, achieves results competitive with fully supervised methodologies, and demonstrates good generalization through external validation~\footnote{Code is available at https://github.com/xmed-lab/CSS-SemiVideo.}. 
    
\end{itemize}

\section{Related Work}
In this section, we review related works on automated LVEF prediction and semi-supervised learning for images and videos.

\subsection{Automated LVEF Prediction}

\textbf{Segmentation-based approach.} LVEF can be predicted from ED and ES frames by estimating EDV and ESV separately from segmentation masks and calculating the percentage difference~\cite{jafari2021deep,reynaud2021ultrasound,wei2020temporal}. Volumes can be directly calculated for 3D segmentation masks and approximated from 2D masks by using method of disks~\cite{bamira2018imaging}. Leclerc \etal~\cite{leclerc2019deep} estimate LVEF by first performing LV segmentation on pre-selected ED and ES frames with a standard U-Net architecture. Liu \etal~\cite{liu2021deep} use pyramid attention modules to learn different feature resolutions for segmentation before calculating LVEF. 

Segmentation approaches for video inputs are more challenging because ED and ES frames must be identified from the input sequence. What's more, LV segmentation labels are typically available only for the ED and ES frame because these are the only labels collected during manual measurement of LVEF~\cite{ouyang2020video,leclerc2019deep} (see Fig. \ref{setting_a}).
Segmentation errors on non-ED and non-ES frames can produce incorrect volume approximations, and these can propagate to the final LVEF value, which is taken as the percentage difference between maximum and minimum volume predictions~\cite{wei2020temporal,painchaud2021echocardiography,yuan2021systematic,chen2022fully}. 
Jafari \etal~\cite{jafari2021deep} use a Bayesian U-Net for frame-wise segmentation and use top and bottom percentile estimates to obtain EDV and ESV. Painchaud \etal~\cite{painchaud2021echocardiography} apply temporal smoothing to volume predictions to reduce outliers. Other works train separate models to first identify ED and ES frames from video before performing segmentation~\cite{dezaki2018cardiac,reynaud2021ultrasound}.

\textbf{Regression-based approach.} State-of-the-art LVEF prediction directly regresses LVEF from video inputs using spatial-temporal models, which avoids the need to estimate EDV and ESV separately. Ouyang \etal~\cite{ouyang2020video} use the R2+1D ResNet~\cite{tran2018closer} to perform regression, achieving an MAE of 4.10. Dai \etal~\cite{dai2021adaptive} further improve predictions by proposing AdaCon, a novel contrastive learning framework for regression problems. 
These end-to-end regression methods produce more reliable estimates by avoiding error propagation from separate frame identification and volume approximation stages. 

Video regression requires large amounts of data for training however. Ouyang and Dai use 10,030 echocardiogram video sequences in their works for example~\cite{ouyang2020video,dai2021adaptive}, which requires extensive labeling effort.
What's more, current regression approaches do not make use of the LV segmentation labels that are collected as part of the labeling process.
Segmentation labels still provide valuable information because LVEF is directly related to changes in the LV. The spatial context from segmentation masks contains location as well as volume information of the LV chamber~\cite{bamira2018imaging}.
Works by Ouyang \etal \cite{ghorbani2020deep} also support the fact that spatial-temporal networks attend to the LV region when learning LVEF regression.
Models trained to perform LV segmentation have specialized knowledge of the LV region, and \textit{this can be used to provide additional guidance for LVEF video regression} if used effectively.

Our proposed method is one of the first to incorporate knowledge from segmentation models into LVEF video regression. By distilling spatial context from LV segmentation predictions into an end-to-end video regression model, we can achieve results competitive with state-of-the-art performance with significantly fewer labeling requirements.

\subsection{Semi-Supervised Learning}

\textbf{Segmentation.} Semi-supervised image segmentation methods combine augmentations with consistency regularization or pseudo-labels to learn from unlabeled data~\cite{tarvainen2017mean,li2018semi,yu2019uncertainty,ouali2020semi,li2020transformation,chen2021semi,yao2022enhancing,lin2022calibrating,wu2021exploring}. The mean-teacher approach~\cite{tarvainen2017mean} uses a teacher model to generate pseudo-labels for the student model. CCT~\cite{ouali2020semi} applies augmentations to intermediate features and enforces consistency on predicted outputs. CPS~\cite{chen2021semi} uses predictions from co-trained segmentation models for cross-supervision. Other approaches using contrastive learning~\cite{alonso2021semi,liu2021bootstrapping} and class activation maps~\cite{zou2020pseudoseg} have also been explored. 
Video segmentation methods makes additional use of temporal information either through optical flow for pseudo-label generation~\cite{yan2019semi} and multi-task learning~\cite{varghese2020unsupervised,ding2020every}, or through multi-frame attention~\cite{perazzi2017learning,oh2019video,cheng2021rethinking}. These methods achieve strong results on benchmark tasks but are typically designed for natural videos. They are not able to utilize the unique characteristics of echocardiogram sequences.

Semi-supervised methods have also been applied to LV segmentation by making use of unlabeled non-ED and non-ES frames in echocardiogram sequences. Works by Zhang \etal~\cite{zhang2020semi} and Pedrosa \etal~\cite{pedrosa2017fast} use different label propagation techniques to generate pseudo-labels for LV segmentation. Others propose jointly learning motion flow and segmentation from video and enforcing consistency between the two tasks~\cite{ta2020semi,qin2018joint,wei2020temporal}.
These methods share similarities with natural video segmentation and do not make use of the special characteristics of echocardiogram sequences. Moreover, they require the ED and ES frames in a sequence to be labeled in order to apply motion propagation and \textit{cannot use completely unlabeled sequences}. 
Our proposed CCS framework uses the cyclical nature of the heart as a prior constraint and can be applied to labeled and unlabeled sequences for improved segmentation and downstream LVEF regression performance (see Table \ref{tab:ablat_seg}).

\textbf{Video Understanding.}
Semi-supervised learning has also been applied to tasks such as video classification, action recognition, and temporal action proposal. State-of-the-art methods share similarities with those for images and use consistency regularization and pseudo-labels on unlabeled data. Xiong \etal~\cite{xiong2021multiview} generate pseudo-labels by ensembling multi-view predictions. Jing \etal~\cite{jing2021videossl} train additional image classifiers to enforce consistency on unlabeled videos for classification. Xu \etal~\cite{xu2021cross} use co-trained models to generate pseudo-labels for each other to perform action recognition. These methods target classification tasks and cannot be applied directly to LVEF regression without modification however. 

Semi-supervised video regression tasks receive less attention compared to classification tasks and have fewer established methods. Ding \etal~\cite{ding2021kfc} propose feature perturbation strategies for semi-supervised temporal action localization. Ji \etal~\cite{ji2019learning} use mean-teacher learning by generating pseudo-labels on unlabeled inputs with a teacher model. These generalized techniques achieve decent performance on natural video but are not well-tailored to echocardiograms. We show in Table~\ref{tab:sota} that our novel approach outperforms existing state-of-the-art methods by effectively making use of the unique properties of the problem domain.

\section{Methodology}

Our proposed methodology consists of two steps: (1) we use a novel cyclical self-supervision (CSS) method to perform semi-supervised LV segmentation using labeled and unlabeled echocardiogram sequences (2) we distill the context provided by LV segmentation predictions into an end-to-end video regression model with teacher-student distillation. Fig.~\ref{flow} illustrates the overall flow. We expand on the two steps in the subsections below. 

We introduce notations and denote $\mathcal{D} \coloneqq \left \{ (X_i, Y_i) \right \}_{i=1}^{N} $ as the labeled dataset with $N$ samples, where $X_i \coloneqq  \{ x_{i}^{1}, x_{i}^{2}, ..., x_{i}^{T_i}  \}$ is an echocardiogram video sequence and $x_{i}^{t}$ is the frame at time-step $t$. 
$Y_i \coloneqq \{ EF_{i},LV_{i}^{ED}, LV_{i}^{ES} \}$ represents the LVEF label and LV segmentation masks for ED and ES frames, respectively. We denote the unlabeled dataset as $\mathcal{D}' \coloneqq \left \{ X'_{i'} \right \}_{i'=1}^{N'} $ which has no corresponding labels. 

The segmentation model consists of a feature encoder and decoder, which we denote $f_{seg} = f_{dec} \circ f_{enc} $. The encoder generates feature embeddings $z_i^t = f_{enc}(x_i^t) \in \mathbb{R}^d$ where $d$ is the feature dimension. The decoder $f_{dec}$ generates LV segmentation predictions based on embeddings, $\hat{LV}_{i}^{t} = f_{dec}(z_{i}^{t})$. 

Two deep regression models are used for LVEF prediction. We denote $f_m$ as the multi-input LVEF regression model that accepts inputs of two types: raw echocardiogram videos concatenated with the LV segmentation mask prediction. 
We denote $f_e$ as the end-to-end model which only accepts raw echocardiogram video inputs.

\subsection{Cyclical Self-Supervision (CSS) for Video-based Semi-Supervised LV Segmentation}

\subsubsection{\wdairv{Motivation}}

The heart pumps blood in a cyclical process and is visually similar during matching phases of the cardiac cycle, such as at the start \wdairv{or end} of contraction. 
\wdairv{This is evident from observing echocardiogram videos as they typically feature repeating sequences of cardiac motion.}
Based on this observation, we aim to train an encoder on labeled and unlabeled echocardiogram sequences by learning feature embeddings that are consistent with the cyclical nature of the heartbeat. Enforcing cyclicality allows us to utilize additional unlabeled data for training, which also helps to reduce over-fitting to labeled ED and ES frames. 

\begin{figure}%
\centering
\begin{subfigure}{\columnwidth}
\centering
\includegraphics[width=0.85 \columnwidth]{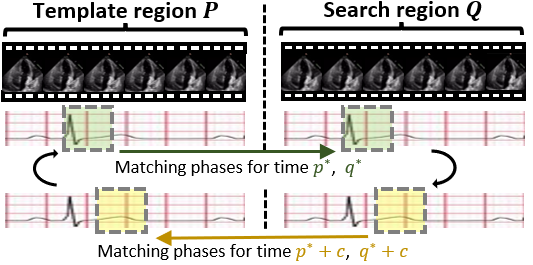}%
\caption{Cyclical relationship within a cardiac sequence}%
\label{detail_a}%
\end{subfigure}\hfill%
\begin{subfigure}{\columnwidth}
\centering
\includegraphics[width=0.85 \columnwidth]{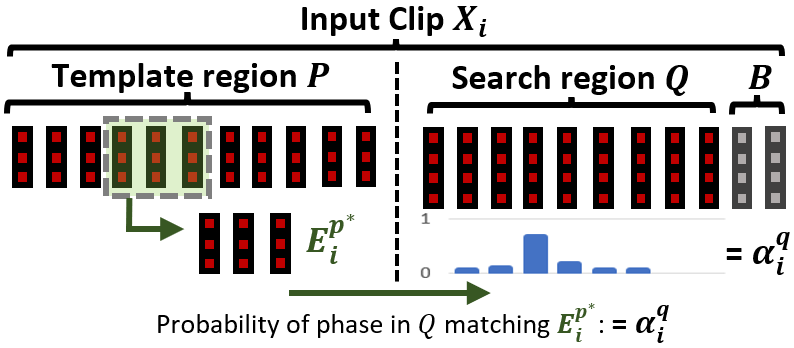}%
\caption{Phase matching in the search region}%
\label{detail_b}%
\end{subfigure}\hfill%
\begin{subfigure}{\columnwidth}
\centering
\includegraphics[width=0.83 \columnwidth]{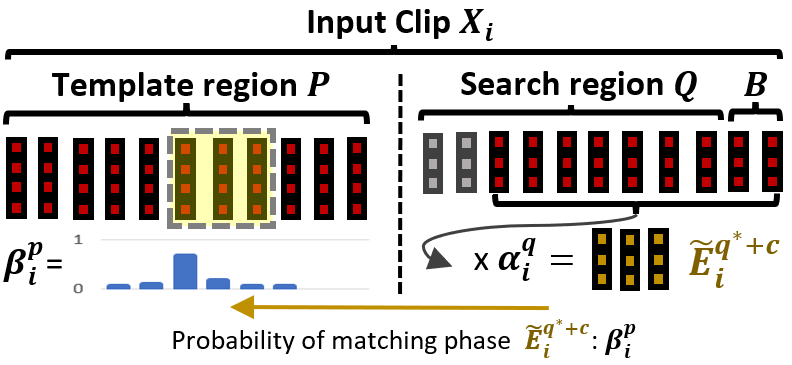}%
\caption{Offset matching in the template region}%
\label{detail_c}%
\end{subfigure}\hfill%

\caption{\textbf{(a)} \wdairv{We tend to perform the following actions when we visually check for cyclicality: 1) We search ahead of our reference point for a matching time-point where the contents are highly similar 2) We skip ahead a number of frames and check that the contents remain aligned. Expressed rigorously: if the phase at time-point $p^*$ of a cyclical process is the same as time-point $q^*$, then the phase at time-point $q^*+c$ is the same as time-point $p^*+c$. 
We enforce this relationship on feature embeddings to reflect cyclicality of cardiac motion.}
\textbf{(b)} Using template phase $E_i^{p^*}$ selected from $P$, we calculate matching probability $\alpha_i^q$ for all time-points $q \in Q$. \wdairv{This is similar to visually searching for a matching phase.} 
\textbf{(c)} We calculate a soft match $\tilde{E}_i^{q^*+c}$ by weighting embeddings by $\alpha_i^q$ and find the closest matching phase in $P$, which should be at $p^*+c$. \wdairv{This is similar to visually checking for continued alignment after a temporal offset of $c$ frames.}  }
\label{detail}
\end{figure}

\wdairv{Identifying and enforcing cyclicality within echocardiograms is challenging however because there are typically no labels associated with cycle duration. The number of frames per cycle varies depending on the patient's heart rate and the video frame rate, and different sequences have different numbers of cycles. To formulate an unsupervised methodology, we observe that humans can recognize cyclical repetition \textit{without cycle labels} by performing the following actions: 1) We first scan ahead of a reference point and find a time-point where the contents are similar to our reference phase; 2) We then skip a few frames ahead of the reference and matching time-points to see if the contents continue to be aligned. If this is true for all sampled references then the sequence is cyclically repeating. 
We can also express this process rigorously through the following relation:
}if a phase at time-point $p^*$ of a cyclical sequence is the same as time-point $q^*$, then the phase at time-point $q^*+c$ is 
the same as time-point $p^*+c$, where $c$ is some temporal offset (see Fig.~\ref{detail_a} for illustration). \wdairv{We design CSS as a differentiable loss function to enforce this relationship on features such that they exhibit temporal cyclicality. 

Labels for $p^*$ and $q^*$ are unavailable for echocardiogram sequences, which means we cannot rely on supervised methods.
Instead, we can use feature similarity to calculate the probability a phase matches with our reference for different time-points. This mimics the process of visually searching for similar content to find a matching phase (see Fig.~\ref{detail_b}). We can then use the probabilities to calculate an expected value for the phase at $q^*+c$ to match with $p^*+c$, which mimics the process of checking for continued alignment (see Fig.~\ref{detail_c}). By enforcing a close match between the two, we ensure the features learned by our model encoder are cyclically consistent. To ensure features are also relevant to the LV segmentation task, we train CSS jointly with supervised segmentation using a shared encoder. We design our CSS loss function based on these considerations. }

\subsubsection{Loss Formulation}
Given some input sequence $X_i$ that covers at least two cardiac cycles, we first divide it into template region $P$, search region $Q$, \wdairv{and a buffer region $B$ used to accommodate frame selection}. 
We analytically \wdairv{define} a phase, \ie a given stage in the cardiac process, by $E^{t}_i$, an array of $s$ feature embeddings starting at time-point $t$:
\begin{equation}
    E_i^{t} = [z_i^{t}, z_i^{t+1}, ..., z_i^{t+s-1}]\:.
\end{equation}
\wdairv{Our design for $E_i^{t}$
captures both spatial and temporal characteristics, which reflects how a person visually identifies video phases based on sequences of key features.
Given some template phase $E_i^{p^*}$ where $p^*$ is randomly selected from $P$, we can enforce cyclical consistency by finding a matching phase in $Q$ and ensuring that the phases continue to match after an equal temporal offset. In other words, we find $q^* \in Q$ and ensure that $E_i^{q^*+c}$ matches with $E_i^{p^*+c}$ for offset $c$.} 

First, we find the probability $\alpha_i^q $  that the phase starting at time-point $q\in Q$ is the same phase as our template $E_i^{p^*}$:

\begin{equation} \label{eq:search_match}
    \alpha_i^q = \frac{{\rm exp}(\gamma(E_i^{p^*},E_i^{q}) \cdot \tau )}{\sum_{j \in Q} {\rm exp}(\gamma(E_i^{p^*},E_i^{j}) \cdot \tau)} \:,
\end{equation}
where $\tau$ is a temperature parameter and $\gamma$ is a similarity function based on inverse squared distance of feature embeddings:
\begin{equation} \label{eq:expeted_match}
\begin{split}
    \gamma(E_i^{p^*},E_i^{j}) & = - \frac{1}{\wdairv{d \times s}} \|E_i^{p^*} - E_i^{j}\|^2 \\
    & = - \frac{1}{\wdairv{d \times s}} \sum_{k=0}^{s-1}  \|z_i^{{p^*}+k} - z_i^{j+k}\|^2  \:. 
\end{split}
\end{equation}
\wdairv{As defined earlier, $d$ is the feature dimension and $s$ is the number of embeddings}. Fig.~\ref{detail_b} illustrates this step, \wdairv{which is similar to visually finding a matching phase}.

Although we do not know the exact value of $E_i^{q^*+c}$ without \wdairv{knowing the true value of} $q^*$, we can calculate a soft expectation for $E_i^{q^*+c}$ by using the matching probabilities $\alpha_i^q$. By finding the probability-weighted average of feature embeddings across time-steps, we can get the expected value of embeddings for $E_i^{q^*+c}$. We define a probability-weighted feature vector $\tilde{z}_i^{\theta}$ as:

\begin{equation}
    \tilde{z}_i^{\theta} = \sum_{j \in Q} \alpha_i^{j} z_i^{j+\theta}   \:.
\end{equation}
where $\theta$ denotes a temporal offset. The expected value for $E_i^{q^*+c}$, which we denote $\tilde{E}_i^{q^*+c}$, can then be formulated as:

\begin{equation}
   \tilde{E}_i^{q^*+c} =  [ \tilde{z}_i^{c} , \tilde{z}_i^{c+1}, ..., \tilde{z}_i^{c+s-1}  ]  \:, 
\end{equation}
\wdairv{where the $c+s-1$ frames that lie outside of region $Q$ are taken from buffer region $B$.}
This soft expectation serves as a differentiable approximation of $E_i^{q^*+c}$ and can be used to perform matching in the template region. We find the probability that the phase at $p \in P $ matches with $\tilde{E}_i^{q^*+c}$ by calculating:
\begin{equation} \label{eq:template_match}
    \beta_i^p = \frac{{\rm exp}(\gamma(\tilde{E}_i^{q^*+c},E_i^{p})  \cdot \tau )}{\sum_{k \in P} {\rm exp}(\gamma(\tilde{E}_i^{q^*+c},E_i^{k})  \cdot \tau)} \:.
\end{equation}
Fig.~\ref{detail_c} illustrates this step\wdairv{, which is similar to visually checking if the contents remain aligned after a temporal offset}. Given the cyclical properties of echocardiograms, $\beta_i^p$ should be highest for $ p^* + c$. We use cross-entropy loss with ground truth $ p^* + c$ to obtain the final CSS loss value $\mathcal{L}_{css}$. Taken across labeled and unlabeled sequences, we have:

\begin{equation}
    \mathcal{L}_{css} = - \frac{1}{N+N'}\sum_{i \in N,N'}\:\sum_{k \in P} \mathbb{1}_{k=p* + c} \log( \beta^{k}_{i})  \:,
\end{equation}
where $N$ and $N'$ refer to the number of labeled and unlabeled samples \wdairv{respectively. 
We note that CSS can be performed even with multiple matching phases present in search region $Q$, which we illustrate in Appendix \ref{apdx_multcyc}. This is vital because it is not possible to guarantee the number of cycles present in the input clip due to variations in cycle rate. CSS does not require any labels regarding cycle length or duration and is the first to perform this task unsupervised.}  

\wdairv{Our objective for enforcing cyclical feature consistency within sequences is to perform semi-supervised segmentation using unlabeled frames. Thus, CSS is trained jointly with supervised LV segmentation through a shared encoder to ensure learned features are relevant to the segmentation task. We obtain predictions on labeled ED and ES frames by applying the segmentation decoder to their feature embeddings:}

\begin{equation}
\hat{LV}_i^{ED}=f_{dec}(z_i^{ED}); \:\: \hat{LV}_i^{ES}=f_{dec}(z_i^{ES}) \: 
\end{equation}
and use the corresponding ground-truth labels $LV_i^{ED}$ and $LV_i^{ES}$ for supervision. 
We apply standard pixel-wise cross-entropy loss to obtain the segmentation loss $\mathcal{L}_{seg}$. The total loss, $\mathcal L_{vol}$, for our semi-supervised LV segmentation method with CSS weight of $w_{css}$ is then:

\begin{equation}
    \mathcal L_{vol} = \mathcal L_{seg} + w_{css} \mathcal L_{css} \: .
\end{equation}
\wdairv{Because CSS is an unsupervised constraint and does not require labels, \textit{sampling for the two tasks is done independently}. 

We also note that CSS is applied on intermediate features instead of segmentation predictions directly to avoid over constraining the model.
This is because motion noise can cause slight changes in the true LV position, and enforcing cyclically equivalent segmentations will force the model to learn wrong predictions. 
By enforcing cyclical consistency only on feature embeddings, which is an abstract representation that captures key characteristics \cite{hu2021semi,ouyang2015learning}, the model is encouraged to learn similar key features for matching phases. Slight variations in LV appearance can then be accurately reflected in predictions by the model decoder.
}

\subsection{Teacher-Student Distillation for LVEF Regression}

LV segmentation masks provide additional context that can supplement LVEF video regression. This is because LVEF is specifically related to changes in the LV and overlaying segmentation masks \textit{gives the regression model additional spatial information about the relevant region}. We first generate LV prediction masks for every frame using our trained segmentation model and concatenate them with raw video as input for multi-input regression. \wdairv{We use a spatial-temporal model for $f_m$ and train it} using available LVEF labels from the labeled dataset with \wdairv{mean squared error} (MSE) loss:

\begin{equation}
    \mathcal L_{m} = \frac{1}{N} \sum_{i \in N} (\hat{EF}_{m,i} - EF_i)^2 \: ,
\end{equation}
where $\hat{EF}_{m,i}$ is the prediction from model $f_m$ for sample $i$. Results in Table~\ref{tab:lab_available} show that LVEF prediction with additional segmentation masks consistently improves performance. 

There are still limitations to the multi-input model however, as inference requires pre-computing segmentation masks for every frame which is impractical. Also, $f_m$ is trained using only labeled data and ignores the unlabeled dataset, which can also provide useful information. We propose teacher-student distillation, which simultaneously performs semi-supervised regression using unlabeled videos, whilst also distilling the multi-input model $f_m$ into an end-to-end model $f_e$. 
\wdairv{Our approach is novel in that \textit{it is the first to distill spatial knowledge from a LV segmentation model into a spatial-temporal video regression model} for LVEF prediction.}

\wdairv{To achieve this, we train $f_e$ on the labeled dataset using LVEF labels and on the unlabeled dataset using pseudo-labels from $f_m$. We use the same architecture as $f_m$ but with a reduced input channel.
Using $f_m$ as a teacher model for pseudo-labels 
serves two purposes. Firstly, $f_e$ learns to predict like the teacher model by using pseudo-labels for guidance, thereby distilling information from segmentation inputs into its own network. Secondly, training with pseudo-labels on unlabeled data is a form of semi-supervised regression, which leads to further improvements over $f_m$ (see ablation results in Table~\ref{tab:lab_available}). The overall result is a more efficient and accurate distilled model, $f_e$. 
We calculate the loss for labeled ($\mathcal{L}_{lb}$) and unlabeled ($\mathcal{L}_{ulb}$) samples using MSE:

\begin{equation}
    \mathcal L_{lb} = \frac{1}{N} \sum_{i \in N} (\hat{EF}_{e,i} - EF_i)^2 \: ,
\end{equation}

\begin{equation}
    \mathcal L_{ulb} = \frac{1}{N'} \sum_{i' \in N'} (\hat{EF}_{e,i'} - \hat{EF}_{m,i'})^2\: .
\end{equation}
where $\hat{EF}_{e,i}$ and $\hat{EF}_{e,i'}$ are the predictions from $f_e$ for sample $i$ and $i'$ respectively.
} 
Training is done by minimizing the loss function $\mathcal L_{dst}$:\wdairv{

\begin{equation}
    \mathcal L_{dst} = \mathcal L_{lb} + w_{ulb} \mathcal L_{ulb}\: ,
\end{equation}
where $w_{ulb}$ is the weight value for unlabeled samples.}
Step 2 of Fig.~\ref{flow} illustrates this step. The final trained model $f_e$ can then be used directly on raw video input for inference without pre-computing segmentation masks.

\section{Experiments}

\begin{table*}[h!]
  \caption{Comparison with state-of-the-art supervised and semi-supervised methodologies using 1/8, 1/4, 1/2, and all training labels \wdairv{on EchoNet-Dynamic}. Note that ``Supervised'' methods use only labeled data while ``Semi-supervised'' methods use both labeled and unlabeled data. \wdairv{We report mean results $\pm$ standard deviation of five separate runs}. 
  }
  \label{tab:sota}
  \centering
  
    \addtolength{\tabcolsep}{-1pt}    
    
    \begin{tabular}{c|c|c|cccc}
    \toprule[1.5pt] 
    
    \multicolumn{7}{c}{MAE Values $\downarrow$}
\\
\hline

    \multirow{1}{*}{Type} & \multirow{1}{*}{Method} & \multirow{1}{*}{Backbone} & 1/8 labels & 1/4 labels & 1/2 labels & All labels \\

\hline

\multirow{2}{*}{Supervised} & Ouyang \etal~\cite{ouyang2020video}      & R2+1D  & 5.64 \wdairv{$\pm$ 0.08} &  4.80  \wdairv{$\pm$ 0.05}  & 4.35 \wdairv{$\pm$ 0.04}  &  4.10 \wdairv{$\pm$ 0.04} \\
 & Dai \etal~\cite{dai2021adaptive} & R2+1D   &  5.47    \wdairv{$\pm$ 0.07} &  4.65  \wdairv{$\pm$ 0.06} &  4.28 \wdairv{$\pm$ 0.04} & \textbf{3.86 \wdairv{$\pm$ 0.03}} \\

\hline

\multirow{3}{*}{\begin{tabular}[c]{@{}c@{}}Semi-\\Supervised\end{tabular}} & 
Ji \etal~\cite{ji2019learning}       & R2+1D  &  5.61 \wdairv{$\pm$ 0.07} & 4.76 \wdairv{$\pm$ 0.06}     &  4.31 \wdairv{$\pm$ 0.04}  & - \\

 & Xu \etal~\cite{xu2021cross}      & R2+1D  &  5.08 \wdairv{$\pm$ 0.04} &  4.49 \wdairv{$\pm$ 0.04}    &  4.28 \wdairv{$\pm$ 0.04}  &  - \\

& Ours    & R2+1D &\textbf{4.90 \wdairv{$\pm$ 0.04}} & \textbf{4.34 \wdairv{$\pm$ 0.03}}    & \textbf{4.17 \wdairv{$\pm$ 0.03}}  & - \\
\hline

\multicolumn{7}{c}{$\mathbf{R}^2$ Values $\uparrow$}
\\
\hline

    \multirow{1}{*}{Type} & \multirow{1}{*}{Method} & \multirow{1}{*}{Backbone} & 1/8 labels & 1/4 labels & 1/2 labels & All labels  \\

\hline

\multirow{2}{*}{Supervised} & Ouyang \etal~\cite{ouyang2020video}      & R2+1D   & 60.6\%  \wdairv{$\pm$ 0.8}   & 71.2\% \wdairv{$\pm$ 0.6} & 76.7\% \wdairv{$\pm$ 0.4}  & 80.5\% \wdairv{$\pm$ 0.2} \\
 & Dai \etal~\cite{dai2021adaptive} & R2+1D    & 62.8\% \wdairv{$\pm$ 0.6}  & 73.4\% \wdairv{$\pm$ 0.5} & 78.1\% \wdairv{$\pm$ 0.3} & \textbf{82.8\% \wdairv{$\pm$ 0.2}} \\

\hline

\multirow{3}{*}{\begin{tabular}[c]{@{}c@{}}Semi-\\Supervised\end{tabular}} & 
Ji \etal~\cite{ji2019learning}       & R2+1D  & 61.6\% \wdairv{$\pm$ 0.6} & 71.4\% \wdairv{$\pm$ 0.6} & 77.2\% \wdairv{$\pm$ 0.4} & -\\

 & Xu \etal~\cite{xu2021cross}      & R2+1D   & 66.8\% \wdairv{$\pm$ 0.4} & 75.8\% \wdairv{$\pm$ 0.3} & 77.6\% \wdairv{$\pm$ 0.3} & -\\

& Ours    & R2+1D  & \textbf{71.1\% \wdairv{$\pm$ 0.4}} & \textbf{77.0\% \wdairv{$\pm$ 0.4}} & \textbf{80.1\% \wdairv{$\pm$ 0.3}} & -\\
\bottomrule[1.5pt]
    
  \end{tabular}
    \addtolength{\tabcolsep}{1pt}    
    
\end{table*}

\subsection{Implementation Details}
\noindent 
\textbf{EchoNet-Dynamic Dataset:} 
We conduct experiments using EchoNet-Dynamic~\cite{ouyang2019echonet}, which consists of 10,030 apical-four-chamber echocardiogram videos labeled with LVEF values and LV segmentation tracings for reference ED and ES frames (Fig.~\ref{setting_a} shows an example). The echocardiogram videos are collected from Stanford University Hospital using five different ultrasound machines with an average length of 175 frames, covering multiple cardiac cycles. The videos are saved using three color channels and have been rescaled to 112$\times$112 pixels with auxiliary text removed. Summary statistics are shown in Table~\ref{lvef_stat}. \wdairv{We also show frame similarity plots in Fig. \ref{frame_sim} of the Appendix to highlight cyclical repetition}. 
The dataset has been \wdairvb{pre-divided} into 7,465 videos for training, 1,288 videos for validation, and 1,277 videos for testing. We \wdairvb{maintain the original data splits but treat} subsets of the training data as labeled data and the remainder as unlabeled data.

\noindent 
\textbf{CAMUS dataset:}
We perform external validation using the four-chamber echocardiogram sequences in CAMUS~\cite{leclerc2019deep}, which are taken from 500 patients at the University Hospital of St. Etienne in France with GE Vivid E95 ultrasound scanners. LVEF labels are available for 450 patients. Sequences begin with the ED frame and terminate with the ES frame or vice-versa. The sequences have an average length of 20 frames and capture only part of the cardiac cycle. They are saved in varying sizes and have been further classified as good, medium, or poor quality \wdairv{as determined by the labeling cardiologist~\cite{leclerc2019deep}}. Summary statistics are shown in Table~\ref{lvef_camus}. 

\noindent 
\textbf{Training CSS:} 
We use the DeepLabV3~\cite{chen2017rethinking} architecture for our segmentation model, which consists of a ResNet-50~\cite{he2016deep} encoder for $f_{enc}$ and an Atrous Spatial Pyramid Pooling (ASPP) segmentation decoder for $f_{dec}$. The model is trained using SGD with a learning rate of $10^{-5}$ and momentum of 0.9. We jointly train the model with CSS and segmentation loss and use mini-batch random sampling for optimization. 

\wdairv{Every iteration, we sample 20 echocardiogram sequences from the labeled dataset and use only their labeled ED and ES frames to minimize $\mathcal{L}_{seg}$. We \textit{separately} sample an input clip of 40 frames from either the labeled or unlabeled dataset at a rate of 1 in every 3 frames to minimize $\mathcal{L}_{css}$. This clip length sufficiently covers two cardiac cycles based on the range of human pulse rates and the frame rate used for EchoNet-Dynamic (see Appendix \ref{apdx_cliplength} for calculations). We set frames 1-15 as template region $P$, 16-36 as search region $Q$, and 37-40 as buffer region $B$. We choose $s=3$ and $c=2$ empirically based on the validation set and set $w_{css} = 0.01$ and $\tau=10$ after performing a coarse parameter search (see Section \ref{css_param}). $p^*$ is randomly sampled every iteration from the range 1-13. 
The frames are passed through the encoder $f_{enc}$ to calculate $\mathcal{L}_{css}$, which is jointly optimized with $\mathcal{L}_{seg}$}. We train for 25 epochs, where an epoch is when all labeled data in $\mathcal{D}$ has been used once for supervised segmentation.

\noindent 
\textbf{Training teacher-student distillation:}
We use the R2+1D ResNet architecture~\cite{tran2018closer}, pretrained on Kinetics 400~\cite{kay2017kinetics}, for $f_m$ and $f_e$. \wdairv{This is the same baseline architecture used by previous works \cite{ouyang2019echonet,ouyang2020video,dai2021adaptive}, although we also include backbone comparisons in Appendix \ref{apdx_backbone}.} The first layer of $f_m$ is modified to accept an additional input channel for multi-input regression. The models are trained with SGD using learning rate $10^{-4}$ and momentum of 0.9.
For minimizing $\mathcal{L}_m$, we sample a batch of 20 video clips taken from labeled echocardiogram sequences for every iteration. The clips are 32 frames in length and sampled at a rate of 1 in every 2 frames. For minimizing $\mathcal{L}_{dst}$, we sample 20 clips from labeled sequences and 10 clips from unlabeled sequences every iteration. \wdairv{We set $w_{ulb} = 5$ after performing a coarse search to determine stable weights (see Section \ref{css_param})}. Both models are trained for 25 epochs. 

We use PyTorch and run experiments on four Titan-RTX GPUs. We perform evaluation using MAE and~$R^2$. 
Each experiment is run five times \wdairvb{using the same data splits with different random seed initializations.} Mean results with standard deviation are reported. 
P-Values where reported are calculated \wdairvb{based on the Student's t-test to indicate level of statistical significance}.

\subsection{Comparison with State-of-the-art Methodologies}\label{sec:sota}

We demonstrate that our proposed method outperforms 
state-of-the-art supervised approaches by Ouyang \etal~\cite{ouyang2020video} and Dai \etal~\cite{dai2021adaptive}, as well as state-of-the-art semi-supervised video learning techniques by Ji \etal~\cite{ji2019learning} and Xu \etal~\cite{xu2021cross} for LVEF prediction. We apply different semi-supervised settings where 1/8, 1/4, and 1/2 of available labels are used for training and the rest is treated as unlabeled data. Supervised methods only use labeled data, whereas semi-supervised methods use both labeled and unlabeled data. We also include results using a fully labeled dataset for supervised methods as reference. 

We directly follow the implementation in~\cite{ouyang2020video,dai2021adaptive} for supervised methods. Because there are no existing semi-supervised video regression approaches used for LVEF prediction, we adapt the methods by Ji \etal~\cite{ji2019learning} and Xu \etal~\cite{xu2021cross} such that they can be applied to our task, which we briefly detail below.  The same backbone architecture, batch size, clip length, and frame sampling rates are used by all methods for fairness. 

\noindent \textbf{Ji \etal~\cite{ji2019learning}:} The authors use a mean-teacher approach together with temporal and frame-wise augmentations for semi-supervised temporal localization. We use the same mean-teacher method but introduce a warm-up period of 10 epochs before using pseudo-labels from the teacher model, which we find leads to better results. We only use random frame jittering and pixel noise as our augmentation operations as we find excessive augmentations handicaps performance. 

\noindent \textbf{Xu \etal~\cite{xu2021cross}:}. The authors propose a cross-model pseudo-labeling scheme, where predictions for weakly augmented inputs are used as pseudo-labels for strongly augmented inputs. We change the classification output to a single regression prediction for our task. We use frame jittering and pixel noise as our strong augmentations and use original frames for weak operations. We use the same R2+1D architectures for the primary and auxiliary backbones to ensure fairness.

We show results of our comparison in Table~\ref{tab:sota}. 
Our method gives the best overall performance across all levels of label availability, achieving MAE of 4.90, 4.34, and 4.17 using one-eighth, one-quarter, and one-half of available labels respectively. \wdairv{$\mathbf{R}^2$ is also significantly higher than alternative approaches for all semi-supervised settings (p-Value~$<$ 0.05).}
Furthermore, we achieve MAE of 4.17 and $\mathbf{R}^2$ of 80.1\% using only half the available labels, which is competitive with MAE of 4.10 and $\mathbf{R}^2$ of 80.5\% achieved by Ouyang \etal~\cite{ouyang2020video} when trained on a fully labeled dataset. 
\begin{figure}%
\captionsetup{labelfont={color=CLRBlue},font={color=CLRBlue}}
\centering
\includegraphics[width=1 \columnwidth]{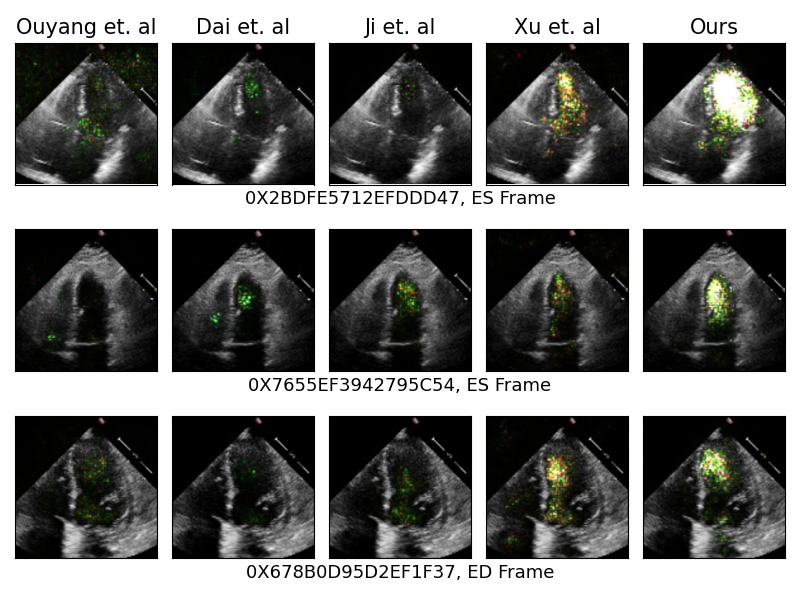}%

\caption{ Qualitative samples of attention heatmaps on EchoNet-Dynamic using different methods. Models are trained using one-eighth of labeled data, with remaining samples treated as unlabeled data. Models attend better to the LV region when trained using our method than alternative approaches.
}
\label{heatmap_echo}
\end{figure}

\begin{table}
\captionsetup{labelfont={color=CLRBlue}}
  \caption{\wdairv{Dice score between LV segmentation labels and pixels with the top 5\% highest absolute gradient values for different methods. Results are shown for the EchoNet-Dynamic dataset. Pixel gradients are calculated using  SmoothGrad~\cite{smilkov2017smoothgrad}. We use models trained with only one-eighth of available labels.}}
\label{tab:dice_echo}
  \centering
  \addtolength{\tabcolsep}{-3.5pt}    
  \color{CLRBlue}
  \begin{tabular}{c|c|cc|c}
    \toprule[1.5pt]
    Type & \multirow{1}{*}{Method} & Dice$\uparrow$ ED  & Dice$\uparrow$ ES  & Dice$\uparrow$ Overall  \\
\hline
\multirow{2}{*}{\begin{tabular}[c]{@{}c@{}}Supervised\end{tabular}} & Ouyang \etal \cite{ouyang2020video}  & 19.2\% & 19.5\%     & 19.4\%    \\
& Dai \etal \cite{dai2021adaptive}  & 25.7\%  & 39.1\%     & 34.0\%     \\
\cline{1-5}

\multirow{3}{*}{\begin{tabular}[c]{@{}c@{}}Semi-\\Supervised\end{tabular}} & Ji \etal \cite{ji2019learning}  & 29.0\%  & 30.0\%     & 29.6\%    \\
& Xu \etal \cite{xu2021cross}  & 31.7\%  & 47.2\%     & 40.9\%     \\
& Ours  & \textbf{41.5\%}   & \textbf{56.1\%}     & \textbf{50.7\%}   \\
\bottomrule[1.5pt]
\end{tabular}
  \addtolength{\tabcolsep}{3.5pt}    

\end{table}

\wdairv{To interpret model performance, we visualize the attention heatmaps generated by the different methods. We use SmoothGrad~\cite{smilkov2017smoothgrad} to calculate the absolute gradient of each input pixel relative to the prediction, which indicates its relative importance to the model, and overlay the normalized values over the frame to generate the heatmap. This is the same approach used by Ouyang \etal \cite{ouyang2020video}.
We use models trained with one-eighth of available labels for comparison. Because our method provides additional spatial context through mask predictions to guide video regression, we expect to see higher absolute gradients for pixels in the LV region compared to alternative approaches. We confirm that this is observed for most samples in the test set and show examples in Fig. \ref{heatmap_echo}. 

To quantify model attention to the LV region, we calculate a Dice score between segmentation labels and the most important pixels, which we determine based on the top 5\% of pixels with the largest absolute gradients. We show results on labeled ED and ES frames in Table \ref{tab:dice_echo}. We can see that the overall Dice value of our method (50.7\%) is significantly higher than alternative methods, indicating greater attention by the model to the LV. Overall, our method is better designed for processing echocardiogram videos and successfully reduces the number of labels required for training. }

\wdairv{

\subsection{Comparing CSS with Unsupervised Temporal Alignment Methods} \label{sec:css_compare_temp}

\begin{table}
\captionsetup{labelfont={color=CLRBlue}}
  \caption{ \wdairv{LVEF prediction results using TCCL compared with CSS (Ours) on EchoNet-Dynamic. One-eighth of labels are used as the labeled dataset, and remaining samples are used as unlabeled data. Results are shown for multi-input model $f_m$ and end-to-end model $f_e$. We report mean results $\pm$ standard deviation of five separate runs.}}
\label{tab:css_vs_tcl}
  \centering
  \addtolength{\tabcolsep}{-3.5pt}    
  \color{CLRBlue}
  \begin{tabular}{c|cc|cc}
    \toprule[1.5pt]
    \multirow{2}{*}{Method} & \multicolumn{2}{c|}{\textbf{$f_m$}} & \multicolumn{2}{c}{\textbf{$f_e$}}\\ \cline{2-5}
  & MAE$\downarrow$  & $\mathbf{R}^2 \uparrow$  & MAE$\downarrow$  & $\mathbf{R}^2 \uparrow$\\
\hline
TCCL~\cite{wei2020temporal}   & 5.25 \wdairv{$\pm$ 0.05} & 65.9\%  \wdairv{$\pm$ 0.6}   & 5.09  \wdairv{$\pm$ 0.04}  & 67.8\%  \wdairv{$\pm$ 0.4} \\
Ours             & \textbf{5.13 \wdairv{$\pm$ 0.05}}  & \textbf{67.6\% \wdairv{$\pm$ 0.5}}      &\textbf{4.90 \wdairv{$\pm$ 0.04}} & \textbf{71.1\% \wdairv{$\pm$ 0.4}} \\
\bottomrule[1.5pt]
\end{tabular}
  \addtolength{\tabcolsep}{3.5pt}    

\end{table}

\begin{table*}[h!]
    \caption{Ablation study\wdairv{ using 3/32, 4/32, 6/32, 8/32, 16/32,} and all training labels \wdairv{on EchoNet-Dynamic}. "Seg." refers to the use of segmentation prediction masks as additional input.
    "CSS" refers to the use of CSS loss for semi-supervised segmentation.
    "Dist." refers to the use of teacher-student distillation. Methods that do not use "Dist." report results from $f_m$. \wdairv{We report mean results $\pm$ standard deviation of five separate runs.}
    }
    \label{tab:lab_available}
  \centering
  
  \addtolength{\tabcolsep}{-1pt}    
  \begin{tabular}{>{\centering\arraybackslash}p{100pt}|>{\centering\arraybackslash}p{10pt}>{\centering\arraybackslash}p{10pt}>{\centering\arraybackslash}p{12pt}|cccccc}
    \toprule[1.5pt] 
    \multicolumn{10}{c}{MAE Values $\downarrow$} \\
    \hline
    \multirow{1}{*}{Method} & \multirow{1}{*}{Seg.} & \multirow{1}{*}{CSS} & \multirow{1}{*}{Dist.} & \wdairv{3/32 labels} & 4/32 labels & \wdairv{6/32 labels} & 8/32 labels & 16/32 labels & All labels \\
\hline
Video Only & & &  & \wdairv{5.82 $\pm$ 0.09} & 5.64 \wdairv{$\pm$ 0.08} & \wdairv{5.22 $\pm$ 0.07}  & 4.80 \wdairv{$\pm$ 0.05} & 4.35 \wdairv{$\pm$ 0.05} & 4.10 \wdairv{$\pm$ 0.04} \\

Video+Seg.  &   \checkmark & &   & \wdairv{5.61 $\pm$ 0.06} & 5.42 \wdairv{$\pm$ 0.07} & \wdairv{5.03 $\pm$ 0.05} & 4.70 \wdairv{$\pm$ 0.05} & 4.29 \wdairv{$\pm$ 0.04} & 4.13 \wdairv{$\pm$ 0.04} \\

Video+Seg.+CSS & \checkmark &  \checkmark  &  & \wdairv{5.47 $\pm$ 0.05} & 5.13 \wdairv{$\pm$ 0.05} &  \wdairv{4.87 $\pm$ 0.03} & 4.67 \wdairv{$\pm$ 0.04} & 4.29 \wdairv{$\pm$ 0.03} & \textbf{4.09 \wdairv{$\pm$ 0.03}} \\

Video+Seg.+Dist. &   \checkmark & &  \checkmark   & \wdairv{5.41 $\pm$ 0.07} & 5.20 \wdairv{$\pm$ 0.06} &  \wdairv{4.72 $\pm$ 0.05} & 4.41 \wdairv{$\pm$ 0.04} & 4.17 \wdairv{$\pm$ 0.03} & -   \\

Video+Seg.+CSS+Dist. (Ours) & \checkmark &  \checkmark  &  \checkmark & \textbf{\wdairv{5.18 $\pm$ 0.06}} &\textbf{4.90 \wdairv{$\pm$ 0.04}} & \textbf{\wdairv{4.55 $\pm$ 0.04}} & \textbf{4.34 \wdairv{$\pm$ 0.03}} & \textbf{4.14 \wdairv{$\pm$ 0.03}} & -\\
\hline

\multicolumn{10}{c}{$\mathbf{R}^2$ Values $\uparrow$} \\
\hline
\multirow{1}{*}{Method} & \multirow{1}{*}{Seg.} & \multirow{1}{*}{CSS} & \multirow{1}{*}{Dist.}  & \wdairv{3/32 labels} & 4/32 labels & \wdairv{6/32 labels} & 8/32 labels & 16/32 labels & All labels \\
\hline
Video Only & & & &  \wdairv{57.2\% $\pm$ 0.9} & 60.6\%  \wdairv{$\pm$ 0.8} & \wdairv{66.9\% $\pm$ 0.6} & 71.2\% \wdairv{$\pm$ 0.6} & 76.7\% \wdairv{$\pm$ 0.4} & \textbf{80.5\% \wdairv{$\pm$ 0.2}}  \\

Video+Seg.  &   \checkmark & &  &  \wdairv{60.4\% $\pm$ 0.7} & 63.0\% \wdairv{$\pm$ 0.7}  & \wdairv{68.7\% $\pm$ 0.6} & 73.5\% \wdairv{$\pm$ 0.3} & 78.3\% \wdairv{$\pm$ 0.4} & 80.3\% \wdairv{$\pm$ 0.3} \\

Video+Seg.+CSS & \checkmark &  \checkmark  &  &  \wdairv{62.6\% $\pm$ 0.6} & 67.6\% \wdairv{$\pm$ 0.5} & \wdairv{71.8\%  $\pm$ 0.3}  & 74.0\% \wdairv{$\pm$ 0.3} & 78.2\% \wdairv{$\pm$ 0.3} & 80.4\% \wdairv{$\pm$ 0.3} \\

Video+Seg.+Dist. &   \checkmark & &  \checkmark  &  \wdairv{62.8\% $\pm$ 0.8}  & 65.2\%  \wdairv{$\pm$ 0.5} &  \wdairv{72.3\% $\pm$ 0.6} & 76.4\% \wdairv{$\pm$ 0.4} & 79.8\% \wdairv{$\pm$ 0.4} & -  \\

Video+Seg.+CSS+Dist. (Ours) & \checkmark &  \checkmark  &  \checkmark &   \textbf{\wdairv{66.3\% $\pm$ 0.6}} & \textbf{71.1\% \wdairv{$\pm$ 0.4}}  & \textbf{\wdairv{74.6\% $\pm$ 0.4}} & \textbf{77.0\% \wdairv{$\pm$ 0.4}} & \textbf{80.1\% \wdairv{$\pm$ 0.3}} & - \\
\bottomrule[1.5pt]
\end{tabular}
\addtolength{\tabcolsep}{1pt}    
\end{table*}

\begin{figure}%
\captionsetup{labelfont={color=CLRBlue},font={color=CLRBlue}}
\centering
\begin{subfigure}{0.45\columnwidth}
\centering
\includegraphics[width=1\columnwidth]{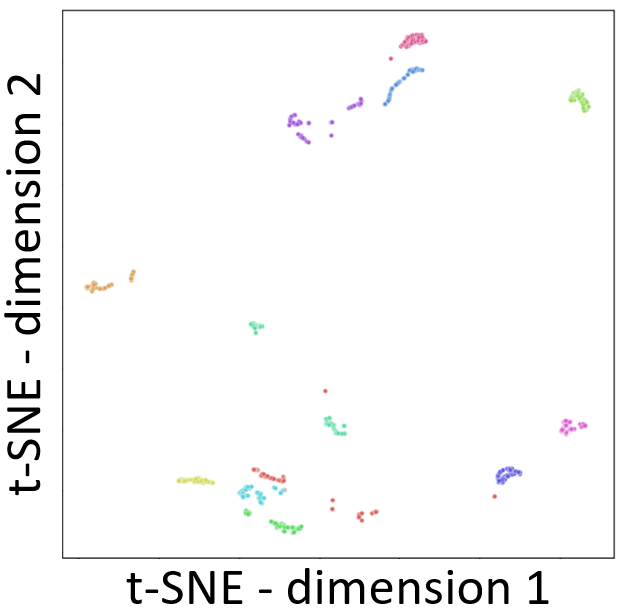}%
\caption{TCCL features}%
\label{tsne_tccl}%
\end{subfigure}\hfill%
\begin{subfigure}{0.45\columnwidth}
\centering
\includegraphics[width=1\columnwidth]{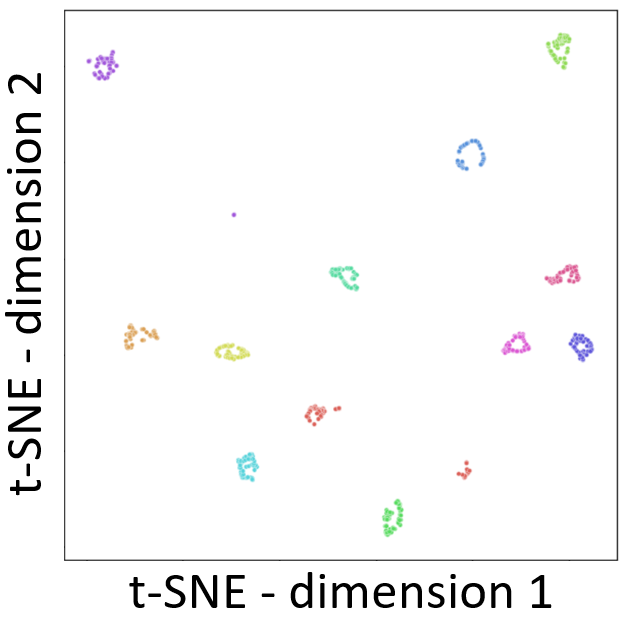}%
\caption{CSS features}%
\label{tsne_css}%
\end{subfigure}\hfill%

\caption{ t-SNE plot of features learned using different methods on EchoNet-Dynamic. Each point represents a single frame of a sequence. Different colors represent different video sequences. 
\textbf{(a)} Features learned using TCCL do not exhibit any cyclicality. Features from different video sequences are also not well separated. 
\textbf{(b)} Features learned using CSS are clustered in circular formations, demonstrating cyclical similarity. Features from different sequences are well separated.  }
\label{feat_tsne}
\end{figure}

\begin{figure*}%
\centering
\begin{subfigure}{\columnwidth}
\centering
\includegraphics[width=0.95 \columnwidth]{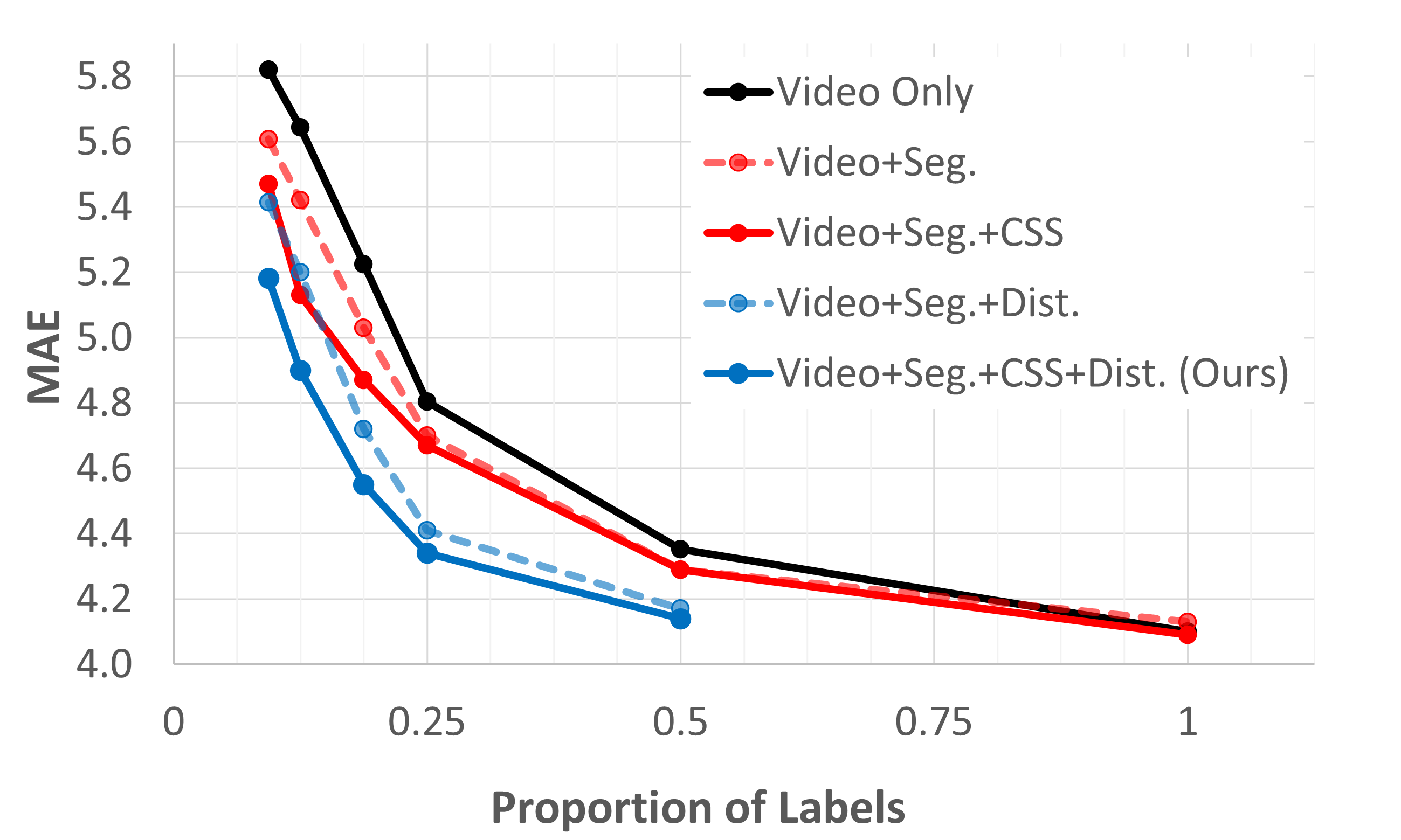}%
\caption{MAE evaluation}%
\label{MAE_datasize}%
\end{subfigure}\hfill%
\begin{subfigure}{\columnwidth}
\centering
\includegraphics[width=0.95 \columnwidth]{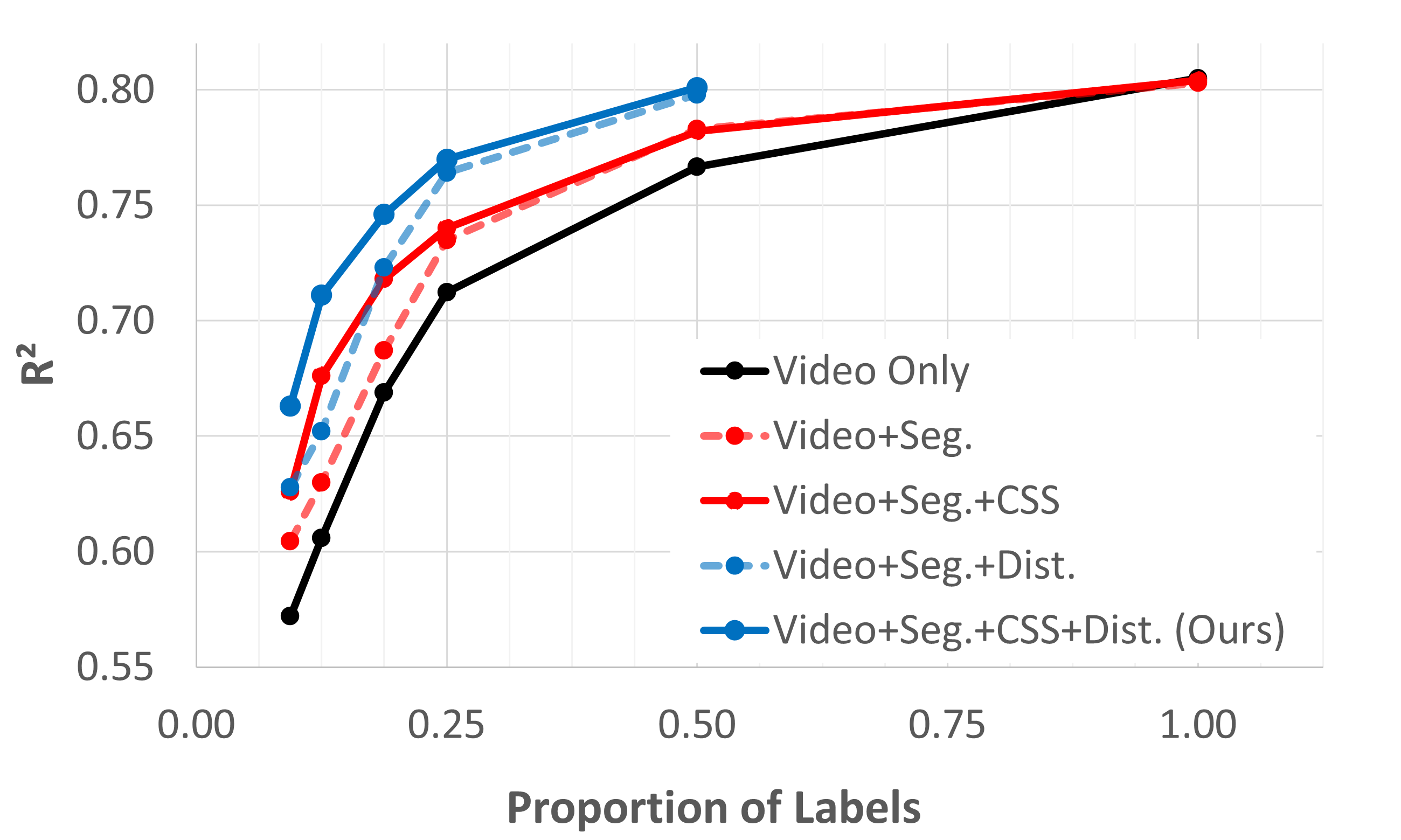}%
\caption{$R^2$ evaluation}%
\label{R2_datasize}%
\end{subfigure}\hfill%

\caption{Plot of Table~\ref{tab:lab_available} showing LVEF prediction results using different proposed components. Training with CSS loss consistently leads to better results compared to without CSS loss (solid lines \vs dashed lines). Training with teacher-student distillation also leads to consistent improvements across different semi-supervised settings (\textcolor{blue}{blue} lines \vs \textcolor{red}{red} lines). 
}
\label{datasize}
\end{figure*}

CSS makes use of cyclical repetition within echocardiograms as a prior for semi-supervised segmentation. We compare with unsupervised feature alignment methods that \textit{do not} enforce cyclical consistency to highlight its effectiveness for LVEF prediction.
One such approach proposed by Dwibedi \etal{} \cite{dwibedi2019temporal} is TCCL, which performs  unsupervised temporal alignment by enforcing feature consistency between the same actions within different videos. Dezaki \etal \cite{dezaki2021echo} also use a similar method to synchronize echocardiograms of different views. Unlike CSS, these methods enforce feature consistency \textit{between different sequences} for temporal alignment instead of \textit{within the same sequence} for ensuring temporal cyclicality. 

We replace CSS with TCCL within our semi-supervised framework and compare results for LVEF prediction. Instead of sampling template region $P$ and search region $Q$ from the same sequence, we sample $Q$ from a separate sequence to perform phase matching and consistency enforcement. This allows the model to learn representations for key phases \textit{across sequences.}
For fairness, we use the same hyper-parameters as our method except for temporal offset $c$, which we do not need for TCCL. We use one-eighth of available labels and treat remaining samples as unlabeled data. Results for multi-input model $f_m$ and end-to-end model $f_e$ are shown in Table~\ref{tab:css_vs_tcl}.

We can see that the MAE of our method is 0.19 lower than TCCL (4.90 \vs{} 5.09, p-Value $<$ 0.05), demonstrating the effectiveness of cyclical consistency. We compare t-SNE plots of segmentation features to further interpret differences between the two methods, which we show in Fig. \ref{feat_tsne}. Each point of the scatter plot represents the feature embedding of a single frame within a sequence. Points with the same colors belong to the same sequence. When using CSS, features from the same video sequence are clustered in circular formations, which is consistent with cyclical repetition. Features from different videos are clearly separated, which also helps with LV segmentation since different sequences have largely different visual appearances.
In contrast, features from different sequences are close together when using TCCL, since key phases are encouraged to have similar features for all sequences. There are also few signs of cyclicality, which means cyclicality of cardiac motion
is not adequately reflected on a feature level. Features learned using CSS are therefore more consistent with our task of semi-supervised segmentation based on cyclical consistency. 

}

\subsection{Ablation Study on Proposed Components}

We analyze the performance contribution from the different proposed components of our method, namely the use of LV segmentations (Seg.), CSS (CSS), and teacher-student distillation (Dist.), by separately combining these modules and comparing their results. This is done using \wdairv{3/32, 4/32, 6/32, 8/32, and 16/32} of available labels to \wdairv{comprehensively show} results under different semi-supervised settings. We also conduct experiments with a fully labeled dataset for methods that do not require unlabeled data for distillation. We briefly detail the combination of modules used in each of the experiments below.
We show results in Table~\ref{tab:lab_available} and visually plot them in Fig.~\ref{datasize}.  

\noindent \textbf{Video Only:} We perform naive regression on raw video input by directly regressing the available LVEF labels. This is the same methodology used by Ouyang \etal~\cite{ouyang2020video}.

\noindent \textbf{Video+Seg.:} We use segmentation masks as additional input for multi-input regression but do not use the proposed CSS and teacher-student distillation methods. A supervised segmentation model is trained using only labeled ED and ES frames and predictions are concatenated with video for input. Only the multi-input model $f_m$ is used. 

\noindent \textbf{Video+Seg.+Dist.:} We perform the same procedure as the ``Video + Seg.'' methodology but use additional teacher-student distillation by introducing $f_e$ and unlabeled sequences.

\noindent \textbf{Video+Seg.+CSS.:} We perform the same procedure as the ``Video+Seg.'' methodology but use CSS loss when training the segmentation model. Teacher-student distillation is not used.

We can see from Fig.~\ref{datasize} that training using only video inputs gives the worst performance (black line). 
Applying CSS loss boosts performance \wdairv{consistently across different settings} (solid lines \vs dashed lines) \wdairv{and significantly decreases MAE (p-Value $<$ 0.05 for 4/32 and 6/32 of labels, p-Value $<$ 0.10 for 3/32 of labels)}. \wdairv{Teacher-student distillation further improves predictions} (blue lines \vs red lines), \wdairv{ especially when used together with CSS (p-Value $<$ 0.05 for MAE reduction). Consistency across different settings provides further evidence of statistical significance. }
Overall, each proposed component has \wdairv{important} contributions to performance.

\subsection{Comparing CSS with Existing Semi-Supervised Segmentation Methods} \label{sec:css_compare}

\wdairv{
\subsubsection{Evaluating performance using LVEF video regression}

We can evaluate the quality of segmentation predictions by concatenating them with video for LVEF video regression.} Higher quality, temporally consistent segmentations provide better spatial guidance to the regression network, which leads to more accurate predictions. \wdairv{This evaluation method is also consistent with the clinical objective of obtaining accurate LVEF measurements through automated methods}. We demonstrate the effectiveness of CSS by comparing with segmentation approaches using CLAS~\cite{wei2020temporal}, multi-frame attention~\cite{ahn2021multi}, and CPS~\cite{chen2021semi}, which represent state-of-the-art methods for semi-supervised image and video LV segmentation. 

The trained models are used to generate LV segmentation mask predictions, which are then concatenated with raw video as input for regression. We use one-eighth of available training labels and treat remaining samples as unlabeled data. A supervised segmentation model trained on labeled data only is also included for comparison. 
We briefly describe implementation details below. 

\noindent \textbf{CLAS~\cite{wei2020temporal}:} Wei \etal simultaneously learn motion flow and LV segmentation from video inputs by enforcing consistency between the two tasks using motion propagation~\cite{wei2020temporal}. We follow the CLAS implementation by Chen \etal~\cite{chen2022fully} which extends the original implementation to EchoNet-Dynamic. 

\noindent \textbf{Multi-frame attention~\cite{ahn2021multi}:} Ahn \etal concatenate feature embeddings from frames surrounding target ED and ES frames to provide additional context for segmentation~\cite{ahn2021multi}. 
We take the three frames before and after the target ED and ES frames as input for the forward and backward directions respectively. 
We use the ResNet-50 encoder~\cite{he2016deep} for $f_{enc}$ and ASPP decoder~\cite{chen2017rethinking} for $f_{dec}$ as our segmentation model and use batch sizes of 20 ED and ES frames for fair comparison. The forward and backward models are trained separately and the average pixel-wise probability is used for prediction.  

\noindent \textbf{CPS~\cite{chen2021semi}:} Chen \etal propose two co-trained segmentation models that generate pseudo-labels for each other~\cite{chen2021semi}. We use the DeepLabV3 architecture~\cite{chen2017rethinking} and jointly train batches of 20 labeled ED and 20 labeled ES frames together with 40 unlabeled frames for every iteration. This uses the same number of labeled and unlabeled frames per iteration as CSS for fairness. 

\noindent \textbf{Supervised:} We train a supervised LV segmentation model using available labels for ED and ES frames. We use DeepLabV3 and a batch size of 20 ED and ES frames for fair comparison.

\begin{table}
  \caption{LVEF prediction results on EchoNet-Dynamic using different state-of-the-art methods to generate segmentation masks for input. One-eighth of labels are used as the labeled dataset, and remaining samples are used as unlabeled data. Results are shown for multi-input model $f_m$ and end-to-end model $f_e$. \wdairv{We report mean results $\pm$ standard deviation of five separate runs.}}
\label{tab:ablat_seg}
  \centering
  \addtolength{\tabcolsep}{-3.5pt}    
  
  \begin{tabular}{c|cc|cc}
    \toprule[1.5pt]
    \multirow{2}{*}{Method} & \multicolumn{2}{c|}{\textbf{$f_m$}} & \multicolumn{2}{c}{\textbf{$f_e$}}\\ \cline{2-5}
  & MAE$\downarrow$  & $\mathbf{R}^2 \uparrow$  & MAE$\downarrow$  & $\mathbf{R}^2 \uparrow$\\
\hline
CLAS~\cite{wei2020temporal}   & 5.72 \wdairv{$\pm$ 0.07} & 59.3\%  \wdairv{$\pm$ 0.9}   & 5.57  \wdairv{$\pm$ 0.07}  & 61.5\%  \wdairv{$\pm$ 0.8} \\
Supervised    & 5.42  \wdairv{$\pm$ 0.07} & 63.0\%  \wdairv{$\pm$ 0.7}    & 5.20  \wdairv{$\pm$ 0.06}   & 65.2\%  \wdairv{$\pm$ 0.5} \\
Multi-frame~\cite{ahn2021multi}  & 5.42 \wdairv{$\pm$ 0.06} & 63.3\% \wdairv{$\pm$ 0.8}  & 5.19 \wdairv{$\pm$ 0.07}  & 66.8\% \wdairv{$\pm$ 0.7}  \\
CPS~\cite{chen2021semi}        & 5.29 \wdairv{$\pm$ 0.06}  & 65.1\% \wdairv{$\pm$ 0.7} & 5.01 \wdairv{$\pm$ 0.04}   & 68.9\% \wdairv{$\pm$ 0.5}   \\
Ours             & \textbf{5.13 \wdairv{$\pm$ 0.05}}  & \textbf{67.6\% \wdairv{$\pm$ 0.5}}      &\textbf{4.90 \wdairv{$\pm$ 0.04}} & \textbf{71.1\% \wdairv{$\pm$ 0.4}} \\
\bottomrule[1.5pt]
\end{tabular}
  \addtolength{\tabcolsep}{3.5pt}    

\end{table}

We show in Table~\ref{tab:ablat_seg} results from multi-input model $f_m$ and distilled model $f_e$. We can see predictions using CSS (Ours) give the lowest MAE, 
achieving 5.13 and 4.90 for $f_m$ and $f_e$ respectively. Improved performance on the multi-input model generally leads to more accurate pseudo-labels, which consequently results in a better distilled end-to-end model.

\begin{table}
\captionsetup{labelfont={color=CLRBlue}}
  \caption{\wdairv{Dice score for LV segmentation predictions on EchoNet-Dynamic. 
  We use only one-eighth of available labels and treat remaining samples as unlabeled data. 
  Results are shown for labeled ED and ES frames and randomly sampled unlabeled frames. 
  We report mean results $\pm$ standard deviation of five separate runs.}
  }
\label{tab:dice_css}
  \centering
  \addtolength{\tabcolsep}{-3.5pt}    
  \color{CLRBlue}
  \begin{tabular}{c|ccc}
    \toprule[1.5pt]
    \multirow{2}{*}{Method} & \multicolumn{3}{c}{Dice$\uparrow$ }\\
    \cline{2-4}
    & ED Frame  & ES Frame  & Unlabeled Frame  \\
\hline
CLAS \cite{wei2020temporal}  & 87.2\% $\pm$ 0.3 & 90.1\% $\pm$ 0.3    & 90.2\% $\pm$ 0.3    \\
Supervised   & 88.8\% $\pm$ 0.3 & 91.4\% $\pm$ 0.2  &  90.8\% $\pm$ 0.2   \\

Multi-frame \cite{ahn2021multi}  & 88.5\% $\pm$ 0.3  & 91.5\% $\pm$ 0.3    & 91.0\% $\pm$ 0.3    \\
CPS \cite{chen2021semi}  & \textbf{89.2\% $\pm$ 0.2}  & 91.9\% $\pm$ 0.2    & 91.3\% $\pm$ 0.2   \\
Ours  & 89.0\% $\pm$ 0.2   & \textbf{92.2\% $\pm$ 0.2}     & \textbf{91.9\% $\pm$ 0.2}   \\
\bottomrule[1.5pt]
\end{tabular}
  \addtolength{\tabcolsep}{3.5pt}    

\end{table}

\subsubsection{Evaluating performance using Dice score}

We can also evaluate the performance of CSS by comparing segmentation prediction results directly. This is challenging however since segmentation labels are only available for reference ED and ES frames, whereas semi-supervised approaches also aim to improve predictions for unlabeled non-ED and non-ES frames \cite{wei2020temporal,qin2018joint}. To address this, we also assess performance on unlabeled frames by randomly sampling a single frame from each of the 1,277 test sequences and manually segmenting the LV. Our segmentation masks were reviewed by a cardiologist to ensure correctness and used as the label for evaluating different segmentation methods. We calculate Dice over ED, ES, and unlabeled frames and present results in Table \ref{tab:dice_css}. We also show qualitative samples in Fig. \ref{unlb_seg_examp} of the Appendix.

We can see from the results that our method, \wdairvb{CSS,} achieves the highest Dice score for ES frames (92.2\%) and unlabeled frames (91.9\%). \wdairvb{CSS outperforms alternatives by a larger margin on unlabeled frames because ED and ES frames are trained with labels by all methods, which leads to similar predictions for these phases. Results are more differentiated on unlabeled frames because different semi-supervision strategies are used.} Results on randomly sampled unlabeled frames are likely to be more indicative of performance on full sequences \textit{since most frames} \textit{are non-ED and non-ES frames}. We also note that the relative performance of different methods is similar to results obtained through LVEF regression. This provides further support that CSS outperforms state-of-the-art alternatives for semi-supervised LV segmentation.

\subsection{Choice of Parameters for CSS}\label{css_param}

\begin{figure}%
\centering
\begin{subfigure}{\columnwidth}
\centering
\includegraphics[width=0.95 \columnwidth]{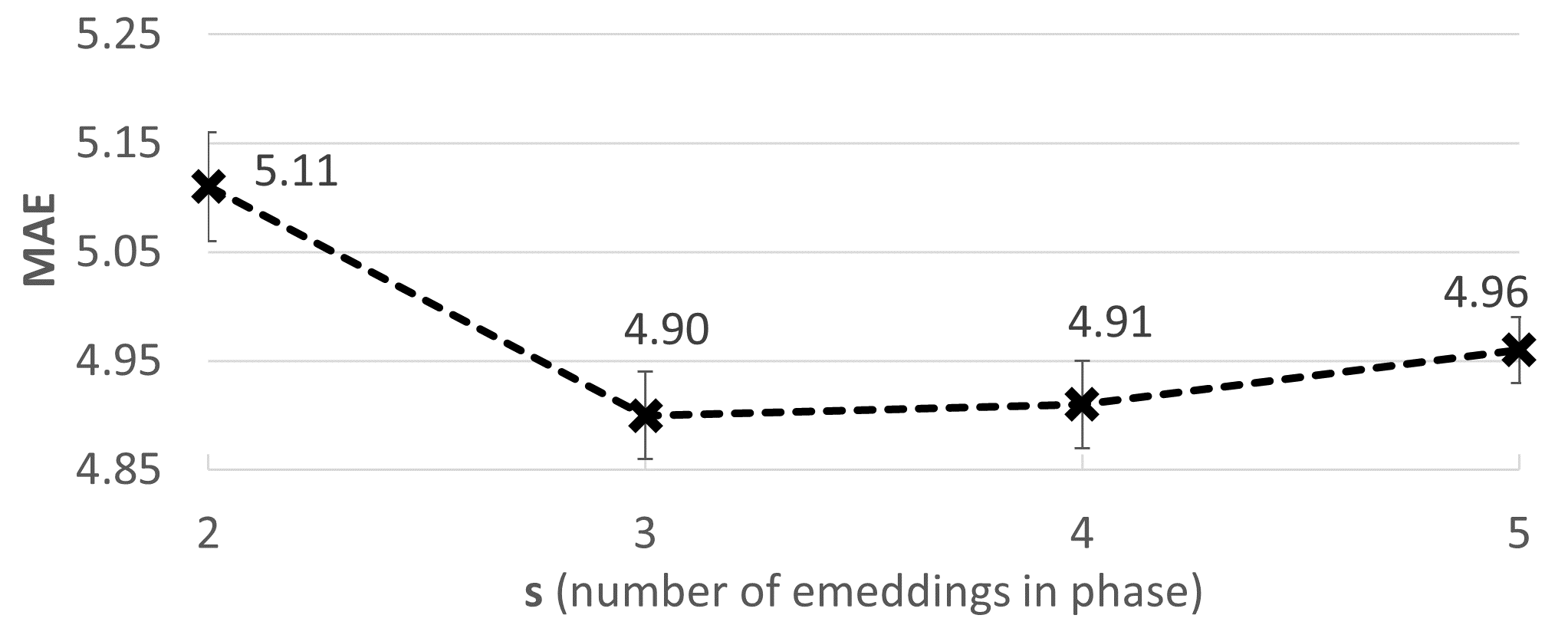}%
\caption{MAE using $s \in \{2,3,4,5\}$, $c=2$, $\tau=10$}%
\label{fig_param_s}%
\end{subfigure}\hfill%

\begin{subfigure}{\columnwidth}
\centering
\includegraphics[width=0.95 \columnwidth]{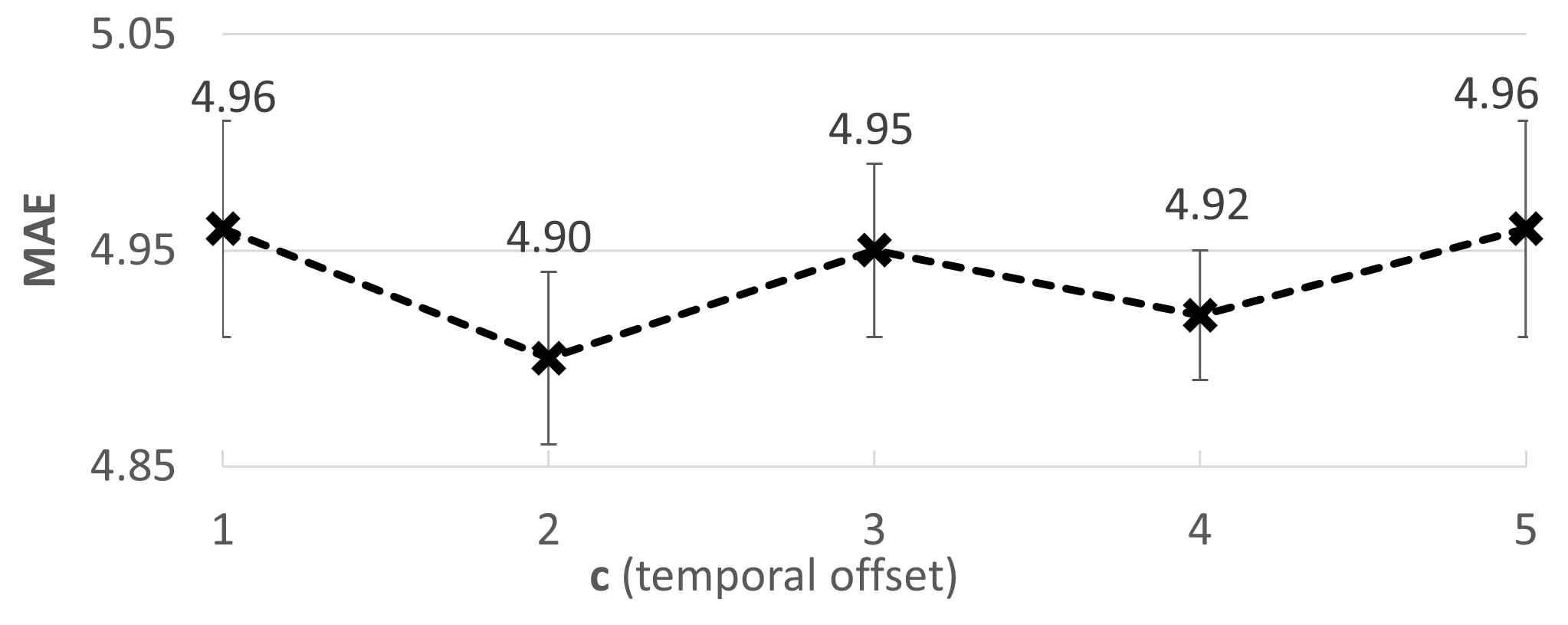}%
\caption{MAE using $c \in \{1,2,3,4,5\}$, $s=3$, $\tau=10$}%
\label{fig_param_c}%
\end{subfigure}\hfill%

\begin{subfigure}{\columnwidth}
\centering
\includegraphics[width=0.95 \columnwidth]{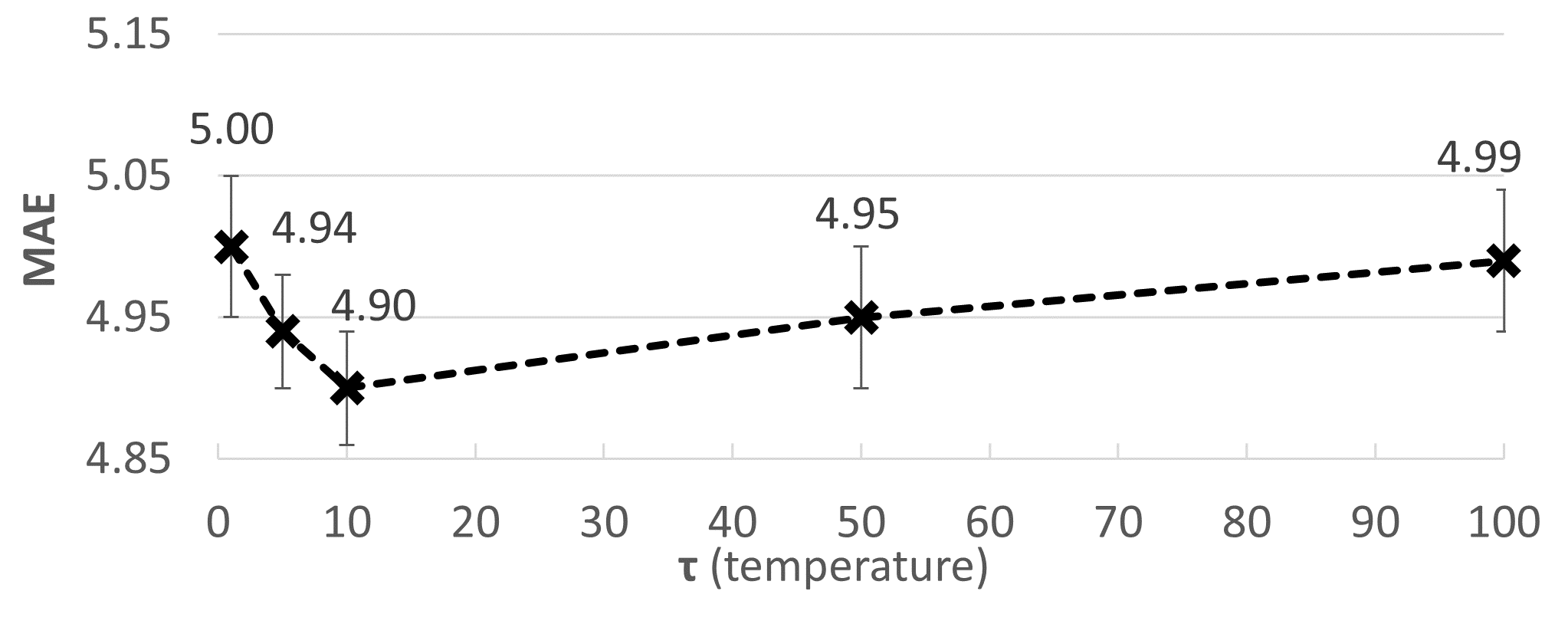}%
\caption{MAE using $\tau \in \{1,5,10,50,100\}$, $s=3$, $c=2$}%
\label{fig_param_t}%
\end{subfigure}\hfill%

\caption{MAE for LVEF prediction on EchoNet-Dynamic using different values of $s$, $c$, and $\tau$ for training CSS. \textbf{(a)} $s$ is the number of feature embeddings used to identify a phase for temporal matching. 
\textbf{(b)} $c$ is the temporal offset used to apply cyclical matching. 
\textbf{(c)} $\tau$ is the softmax temperature parameter controlling the sharpness of the matching probability profile. \wdairv{We use one-eighth of labeled data as the labeled dataset and treat remaining samples as unlabeled data.} 
}
\label{fig_param}
\end{figure}

The parameters used for training CSS include the number of feature embeddings in a phase ($s$), the temporal offset ($c$), and the softmax temperature ($\tau$). We test the sensitivity of our method to parameter choice by varying parameter values one at a time whilst keeping others constant at our chosen settings $s=3$, $c=2$, and $\tau=10$. We \wdairv{perform experiments using one-eighth of available training labels as the labeled dataset and treat the remaining as unlabeled data. Results} using values $s \in \{2,3,4,5\}$, $c \in \{1,2,3,4,5\}$ , and  $\tau \in \{1,5,10,50,100\}$ \wdairv{are visually plotted} in Fig. \ref{fig_param}.

We can see from Fig. \ref{fig_param_s} that results are generally stable \wdairv{between 4.90 and 4.96 MAE} for $s >= 3$. Results using $s=2$ are slightly weaker as fewer feature embeddings are used for phase identification, which may lead to incorrect phase matching. Fig. \ref{fig_param_c} shows that results for different values of $c$ are stable between 4.90 and 4.96 MAE. Fig. \ref{fig_param_t} shows that $\tau=10$ is the most suitable value for temperature.

\wdairv{We also perform a coarse search for parameters $w_{css} \in \{0.005,0.01,0.02\}$ and $w_{ulb} \in \{2,5,10\}$ using one-eighth of available training labels to determine stable training weights. For $w_{css}$, we show results from the multi-input model $f_m$ and the distilled model $f_e$ in Table \ref{tab:result_wcss}. We can see using $w_{ulb}=0.01$ gives us the best results, although performance is relatively robust to different values. Results for $w_{ulb}$ are shown in Table \ref{tab:result_wulb}, where we can see that using $w_{ulb}=5$ gives us the best results.}

\begin{table}
\captionsetup{labelfont={color=CLRBlue}}

  \caption{\wdairv{LVEF prediction results using different values for $w_{css}$ on EchoNet-Dynamic. One-eighth of labels are used as the labeled dataset, and remaining samples are used as unlabeled data. Results are shown for multi-input model $f_m$ and end-to-end model $f_e$. We report mean results $\pm$ standard deviation of five separate runs.}}
\label{tab:result_wcss}
  \centering
  
  \addtolength{\tabcolsep}{-3.1pt}  
  \color{CLRBlue}
  \begin{tabular}{c|cc|cc}
    \toprule[1.5pt]
\multirow{2}{*}{Method} & \multicolumn{2}{c|}{\textbf{$f_m$}} & \multicolumn{2}{c}{\textbf{$f_e$}}\\ \cline{2-5}
  & MAE$\downarrow$  & $\mathbf{R}^2 \uparrow$  & MAE$\downarrow$  & $\mathbf{R}^2 \uparrow$\\
\hline
$w_{css} = 0.02$    & 5.24 $\pm$ 0.07 & 66.4\% $\pm$ 0.6  & 4.99 $\pm$ 0.06 & 70.1\% $\pm$ 0.6 \\
$w_{css} = 0.01$             & \textbf{5.13 $\pm$ 0.05}  & \textbf{67.6\% $\pm$ 0.5}      &\textbf{4.90 $\pm$ 0.04} & \textbf{71.1\% $\pm$ 0.4} \\
$w_{css} = 0.005$    & 5.19 $\pm$ 0.06 & 66.9\% $\pm$ 0.5   & 4.96 $\pm$ 0.05 & 70.3\% $\pm$ 0.5 \\
\bottomrule[1.5pt]
\end{tabular}

  \addtolength{\tabcolsep}{3.1pt}  
\end{table}

\begin{table}
\captionsetup{labelfont={color=CLRBlue}}
  \caption{\wdairv{LVEF prediction results using different values for $w_{ulb}$ on EchoNet-Dynamic. One-eighth of labels are used as the labeled dataset, and remaining samples are used as unlabeled data. We report mean results $\pm$ standard deviation of five separate runs.}}
\label{tab:result_wulb}
  \centering
  \color{CLRBlue}
  \begin{tabular}{c|cc}
    \toprule[1.5pt]
    Method &  MAE$\downarrow$  & $\mathbf{R}^2 \uparrow$  \\
\hline
$w_{ulb}=10$   &  5.13 $\pm$ 0.04 & 68.2\%    $\pm$ 0.4 \\
$w_{ulb}=5$     & \textbf{4.90 $\pm$ 0.04} & \textbf{71.1\% $\pm$ 0.4}   \\
$w_{ulb}=2$   & 4.96 $\pm$ 0.05 & 70.1\% $\pm$ 0.4\\
\bottomrule[1.5pt]
\end{tabular}

\end{table}

\begin{table*}[h!]
    \caption{External validation of trained models on the CAMUS dataset. 
    MAE is reported based on video quality and overall performance. Models are trained using 1/8, 1/4, 1/2, and all of the available training labels on the EchoNet-Dynamic dataset and tested on CAMUS. Note that ``Supervised'' methods use only labeled data while ``Semi-supervised'' methods use both labeled and unlabeled data. \wdairv{We show mean results $\pm$ standard deviation of the five training runs.} }
    \label{tab:external_val}
  \centering

\addtolength{\tabcolsep}{-1pt}    
  \begin{tabular}{c|c|c|cccc} 
    
    \toprule[1.5pt]
    \multicolumn{7}{c}{MAE Values $\downarrow$} \\
    \hline
    \multirow{1}{*}{Quality} &\multirow{1}{*}{Type} & \multirow{1}{*}{Method} & 1/8 labels & 1/4 labels & 1/2 labels & All labels \\ 
\hline

\multirow{5}{*}{Good} & \multirow{2}{*}{Supervised}  & Ouyang \etal~\cite{ouyang2020video} & 7.27 \wdairv{$\pm$ 0.19} & 6.43 \wdairv{$\pm$ 0.14} & 6.37 \wdairv{$\pm$ 0.11} & 6.21 \wdairv{$\pm$ 0.08} \\ 

  & & Dai \etal~\cite{dai2021adaptive} & 7.24 \wdairv{$\pm$ 0.18} & 6.40 \wdairv{$\pm$ 0.13} & 6.31 \wdairv{$\pm$ 0.10} & \textbf{5.90 \wdairv{$\pm$ 0.07}} \\ 
\cline{2-7}

  & \multirow{3}{*}{\begin{tabular}[c]{@{}c@{}}Semi-\\Supervised\end{tabular}} & Ji \etal~\cite{ji2019learning} & 7.87 \wdairv{$\pm$ 0.18} & 6.49 \wdairv{$\pm$ 0.13} & 6.10 \wdairv{$\pm$ 0.10} & - \\ 

  & & Xu \etal~\cite{xu2021cross}  & 6.63 \wdairv{$\pm$ 0.12} & 6.39 \wdairv{$\pm$ 0.13} & 5.95 \wdairv{$\pm$ 0.12} & - \\ 

  &  & Ours & \textbf{6.51 \wdairv{$\pm$ 0.12}} & \textbf{6.01 \wdairv{$\pm$ 0.11}}& \textbf{5.67 \wdairv{$\pm$ 0.10}} & - \\ 
  
\hline
\hline

\multirow{5}{*}{Medium} & \multirow{2}{*}{Supervised} & Ouyang \etal~\cite{ouyang2020video} & 8.61 \wdairv{$\pm$ 0.24} & 7.96 \wdairv{$\pm$ 0.14} & 8.53 \wdairv{$\pm$ 0.15} & 6.89 \wdairv{$\pm$ 0.13} \\ 

  & & Dai \etal~\cite{dai2021adaptive} & 8.85 \wdairv{$\pm$ 0.22} & 7.49 \wdairv{$\pm$ 0.16} & 7.77 \wdairv{$\pm$ 0.15} &\textbf{6.72 \wdairv{$\pm$ 0.10}} \\ 
    \cline{2-7}
    
  & \multirow{3}{*}{\begin{tabular}[c]{@{}c@{}}Semi-\\Supervised\end{tabular}} & Ji \etal~\cite{ji2019learning} & 8.54 \wdairv{$\pm$ 0.18} & 7.38 \wdairv{$\pm$ 0.17} & 7.34 \wdairv{$\pm$ 0.12} & - \\ 

  &  & Xu \etal~\cite{xu2021cross}  &  8.41 \wdairv{$\pm$ 0.14} & 7.52 \wdairv{$\pm$ 0.14} & 7.19 \wdairv{$\pm$ 0.15} & - \\ 
  
  & & Ours  & \textbf{8.33 \wdairv{$\pm$ 0.13}} & \textbf{7.20 \wdairv{$\pm$ 0.10}} & \textbf{6.34 \wdairv{$\pm$ 0.15}} & - \\ 
  
\hline
\hline

\multirow{5}{*}{Poor} & \multirow{2}{*}{Supervised}  & Ouyang \etal~\cite{ouyang2020video} & 11.00 \wdairv{$\pm$ 0.26} & 9.43 \wdairv{$\pm$ 0.19} & 10.64 \wdairv{$\pm$ 0.15} & \textbf{10.33 \wdairv{$\pm$ 0.15}} \\ 

  & & Dai \etal~\cite{dai2021adaptive} & 11.31 \wdairv{$\pm$ 0.26} & 9.53 \wdairv{$\pm$ 0.20} & 10.88 \wdairv{$\pm$ 0.14} & 10.82 \wdairv{$\pm$ 0.11} \\ 
    \cline{2-7}
    
  & \multirow{3}{*}{\begin{tabular}[c]{@{}c@{}}Semi-\\Supervised\end{tabular}} & Ji \etal~\cite{ji2019learning} & \textbf{9.71 \wdairv{$\pm$ 0.25}}& 8.95 \wdairv{$\pm$ 0.22} & 8.51 \wdairv{$\pm$ 0.14} & - \\ 

  &  & Xu \etal~\cite{xu2021cross}  & 10.36 \wdairv{$\pm$ 0.18} & 9.32 \wdairv{$\pm$ 0.16} & 9.51 \wdairv{$\pm$ 0.14} & - \\ 
  
  & & Ours & 10.77 \wdairv{$\pm$ 0.13} & \textbf{8.78 \wdairv{$\pm$ 0.15}} & \textbf{8.00 \wdairv{$\pm$ 0.16}} & - \\ 
  
\hline
\hline
\multirow{5}{*}{Overall} & \multirow{2}{*}{Supervised} & Ouyang \etal~\cite{ouyang2020video} & 8.06 \wdairv{$\pm$ 0.22} & 7.21 \wdairv{$\pm$ 0.17} & 7.48 \wdairv{$\pm$ 0.14} & 6.82 \wdairv{$\pm$ 0.12} \\ 

  & & Dai \etal~\cite{dai2021adaptive} & 8.15 \wdairv{$\pm$ 0.18} & 7.05 \wdairv{$\pm$ 0.16} & 7.22 \wdairv{$\pm$ 0.13} & \textbf{6.63 \wdairv{$\pm$ 0.06}} \\ 
 \cline{2-7}
  & \multirow{3}{*}{\begin{tabular}[c]{@{}c@{}}Semi-\\Supervised\end{tabular}} & Ji \etal~\cite{ji2019learning} & 8.26 \wdairv{$\pm$ 0.20} & 7.01 \wdairv{$\pm$ 0.14} & 6.73 \wdairv{$\pm$ 0.13} & - \\ 

  &  & Xu \etal~\cite{xu2021cross}  & 7.57 \wdairv{$\pm$ 0.15} & 7.04 \wdairv{$\pm$ 0.16} & 6.69 \wdairv{$\pm$ 0.11} & - \\ 
  
  & & Ours & \textbf{7.51 \wdairv{$\pm$ 0.14}} & \textbf{6.66 \wdairv{$\pm$ 0.09}} & \textbf{6.11 \wdairv{$\pm$ 0.12}} & - \\ 

\bottomrule[1.5pt]
\end{tabular}
\addtolength{\tabcolsep}{1pt}    
\end{table*}

\subsection{External Validation of Trained Models}

Model generalization is desirable for medical applications as different hospitals may have different equipment and data collection procedures, leading to domain shifts. We perform external validation of the models trained in Section~\ref{sec:sota} using the four-chamber sequences from the CAMUS dataset~\cite{leclerc2019deep} to test for generalized performance. CAMUS sequences have significantly fewer frames on average compared to EchoNet-Dynamic (20 \vs 175) and only capture part of the cardiac cycle. In contrast, sequences for EchoNet-Dynamic cover multiple cycles and to the best of our knowledge is the only publicly available dataset with multi-cycle data. For pre-processing, we double the input sequence by mirroring the data along the temporal dimension at the start and end of the sequence to simulate a full cardiac cycle \wdairvb{(see Appendix \ref{apdx_mirror} for details)}. \wdairv{Temporal mirroring helps preserve smoothness since the sequence is simply reversed and no discontinuity is introduced. We simulate a full cardiac cycle for testing since this is more consistent with echocardiograms obtained through realistic acquisition procedures \cite{leeson2012echocardiography}. Frames are resized to 112$\times$112 pixels to fit the trained model.} Inference results are shown in Table~\ref{tab:external_val}. 

\begin{figure}%
\captionsetup{labelfont={color=CLRBlue},font={color=CLRBlue}}
\centering
\includegraphics[width=0.92 \columnwidth]{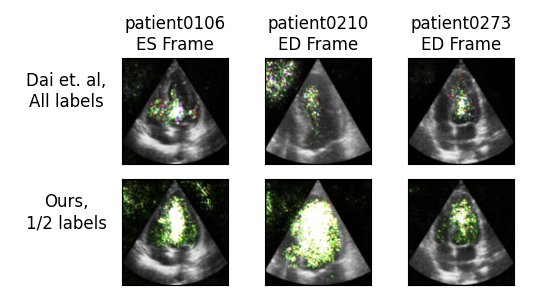}%

\caption{Qualitative samples of attention heatmaps on CAMUS. Models are trained using EchoNet-Dynamic under different methods and semi-supervised settings. Models trained using our method on half of available labels attend better to the LV region than Dai \etal \cite{dai2021adaptive} trained on full labels. }
\label{heatmap_CAMUS}
\end{figure}

We can see our proposed method consistently outperforms alternatives for good and medium-quality sequences and has the best overall performance. 
Our method trained using half the available labels on EchoNet-Dynamic also outperforms the supervised method by Dai \etal~\cite{dai2021adaptive} trained on full labels\wdairv{, demonstrating superior generalization   (MAE 6.11~\vs 6.63 with p-Value $<$ 0.05)}. \wdairv{To understand why this is the case, we compare attention heatmaps to visualize the regions that are important to the models for inference.
We use SmoothGrad \cite{smilkov2017smoothgrad} to calculate pixel gradient contributions, where higher absolute pixel gradients indicate greater sensitivity to changes in that region. For a well generalized model, we expect pixels in the LV to have higher absolute gradients since LVEF is primarily determined by changes in this region. We confirm that our method generates heatmaps that are more concentrated in the LV region compared to Dai \etal{} \cite{dai2021adaptive} and show qualitative examples in Fig. \ref{heatmap_CAMUS}. 

To quantify attention to the LV region, we calculate the degree of overlap between segmentation labels and the most important pixels. We calculate the Dice score between LV mask labels and pixels with the top 5\% highest absolute gradient values and show results in Table \ref{tab:dice_camus}. We can see our method, even when trained using half the available labels, has a higher Dice score compared to Dai \etal \cite{dai2021adaptive} trained on a full dataset. This shows that our model can identify important regions more effectively compared to supervised approaches, even when tested on external datasets.
Existing works have also shown that training with unlabeled data leads to better model generalization since it reduces over-fitting to the labeled dataset \cite{zhong2022self,hendrycks2019using}. This further supports the use of our proposed semi-supervised method for LVEF regression.}

\begin{table}
    \captionsetup{labelfont={color=CLRBlue}}
  \caption{\wdairv{Dice score between LV segmentation labels and pixels with the top 5\% highest absolute gradient values. Results are shown for the CAMUS dataset with models trained on EchoNet-Dynamic using different methods and different semi-supervised settings. Pixel gradients are calculated using SmoothGrad~\cite{smilkov2017smoothgrad}.}}
\label{tab:dice_camus}
  \centering
  \addtolength{\tabcolsep}{-3.5pt}    
  
  \color{CLRBlue}
  \begin{tabular}{c|c|cc|c}
    \toprule[1.5pt]
     \multirow{1}{*}{Method} & Setting & Dice$\uparrow$ ED  & Dice$\uparrow$ ES  & Dice$\uparrow$ Overall  \\
\hline

 Dai \etal \cite{dai2021adaptive} & All labels & 42.9\%  & 47.6\%     & 45.4\%     \\
 Ours & 1/2 labels  & \textbf{46.7\%}   & \textbf{57.1\%}     & \textbf{52.3\%}   \\
\bottomrule[1.5pt]
\end{tabular}
  \addtolength{\tabcolsep}{3.5pt}    

\end{table}

\wdairv{
\section{Discussion}

\subsection{Impact of Heart Arrhythmia on CSS}

Heart arrhythmia refers to when the heart either beats too fast, too slow, or at an irregular pace with altering rates~\cite{tse2016mechanisms}. Our CSS method was designed with heart arrhythmia in consideration since cases had previously been identified in EchoNet-Dynamic by other studies \cite{ouyang2020video}. CSS does not assume consistent periodic cycles across the entire video sequence and only enforces consistency between adjacent cycles of sampled clips. Because of this, it is relatively robust to irregular heart rates. 
Although a more detailed study on cardiac arrhythmia cases will be informative, this would require expert labels which are currently unavailable. We leave this to future investigation and note that current experimental results validate the effectiveness of our method even with such cases present in the dataset. 

\subsection{Choice of temporal offset}

We use a constant temporal offset of $c=2$ in our method, which means we apply an offset of two frames from the reference and matching time-points to check for cyclical consistency. Similar results were also achieved using alternative values. 
Another possible choice is applying an offset value that is relative to cycle length, since different patients have different cycle durations. Doing so will introduce additional complexities however, since labels for cycle lengths are unavailable and can only be inferred indirectly from feature matching probabilities. Furthermore, unlike a constant temporal offset, which can be applied directly to all samples, a relative offset may lead to fractional values and may require temporal interpolation between frames for implementation. Given that our method already outperforms alternative approaches and \textit{is the first to make use of unsupervised cyclical consistency}, we do not explore relative offsets at this time.
}

\section{Conclusion}

In this work, we introduce the first semi-supervised approach to LVEF prediction from echocardiogram videos. We address the problem in two steps by proposing a novel cyclical self-supervision (CSS) method for semi-supervised video LV segmentation, and a teacher-student distillation method for distilling spatial context from segmentation predictions into an end-to-end video regression model.
Our method allows echocardiogram videos to be trained using fewer labels, which is important for reducing labeling costs. Experiments show our method outperforms existing semi-supervised alternatives and achieves performance competitive with supervised methods on fully labeled datasets. Furthermore, external validation demonstrates that our method has better generalization when tested on data from a different hospital.

\appendices
\setcounter{table}{0}
\renewcommand{\thetable}{A\arabic{table}}
\setcounter{figure}{0}
\renewcommand{\thefigure}{A\arabic{figure}}
\setcounter{equation}{13}

\wdairv{

\begin{figure}%
\captionsetup{labelfont={color=CLRBlue},font={color=CLRBlue}}
\centering
\begin{subfigure}{\columnwidth}
\centering
\includegraphics[width=0.92 \columnwidth]{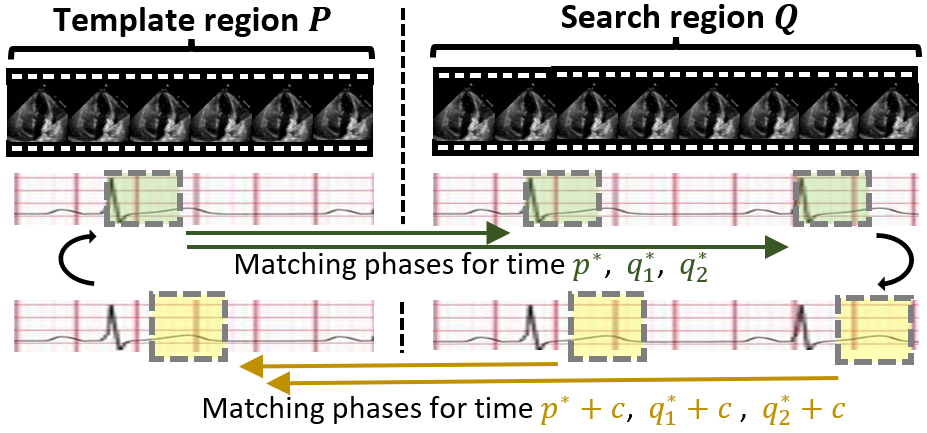}%
\caption{Cyclical relationship within a sequence for multiple cycles}%
\label{detail_a_mult}%
\end{subfigure}\hfill%
\begin{subfigure}{\columnwidth}
\centering
\includegraphics[width=0.92 \columnwidth]{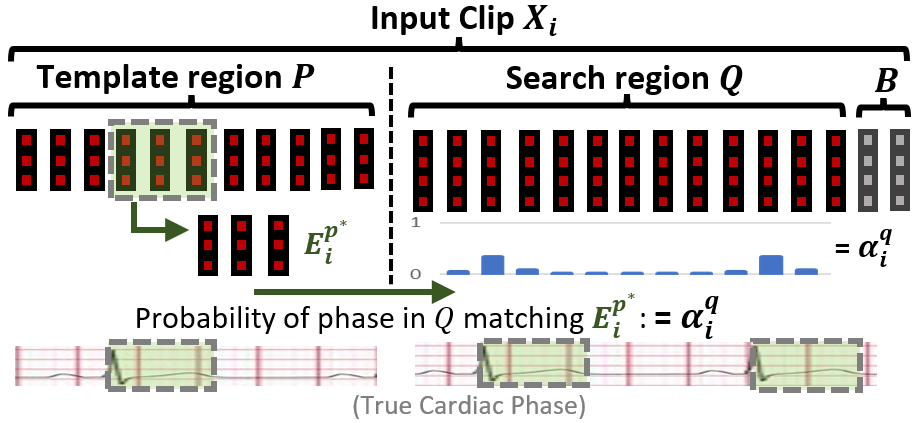}%
\caption{Phase matching in the search region with two cycles present}%
\label{detail_b_mult}%
\end{subfigure}\hfill%
\begin{subfigure}{\columnwidth}
\centering
\includegraphics[width=0.92 \columnwidth]{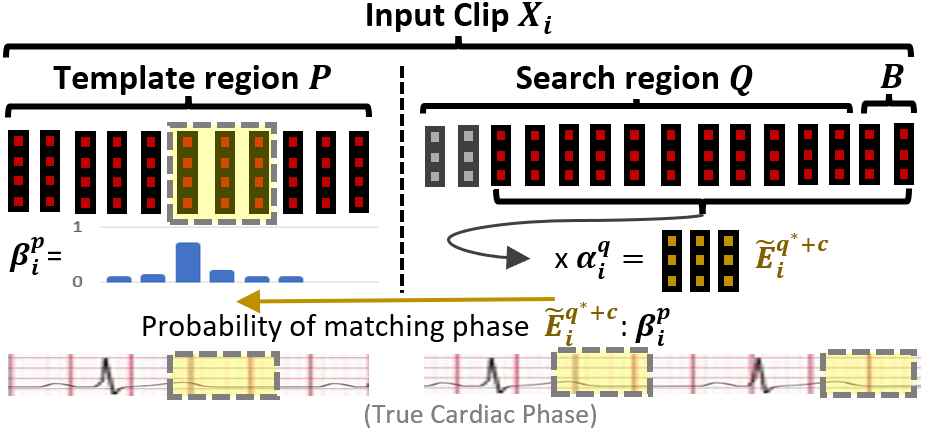}%
\caption{Offset matching in the template region}%
\label{detail_c_mult}%
\end{subfigure}\hfill%

\caption{\textbf{(a)} We can enforce the same relationship for cyclicality even if three cycles are in the input sequence: if the phase at time-point $p^*$ is the same as time-point $q^*_1$ and $q^*_2$ for a cyclical sequence, then the phase at time-point $q^*_1+c$ and $q^*_2+c$ is the same as time-point $p^*+c$
\textbf{(b)} Using template phase $E_i^{p^*}$, we calculate the matching probability, $\alpha_i^q$, in the search region for $q \in Q$. For well trained features, the time-points corresponding to true matches will have higher probabilities. 
\textbf{(c)} $\tilde{E}_i^{q^*+c}$ can still be calculated based on matching probabilities even with multiple cycles. We weigh embeddings by $\alpha_i^q$ and find the closest matching time-point in $P$, which should be $p^*+c$. The same loss function can still be used.  }
\label{detail_mult}
\end{figure}

\section{Cyclical feature matching with multiple cardiac cycles in the search region} \label{apdx_multcyc}

One of the challenges in enforcing cyclical feature consistency is that the number of frames per cycle within a video sequence can vary. This means it is impossible to guarantee the number of cycles present in input clip $X_i$. 
The illustrations and examples in Fig. \ref{flow} and Fig. \ref{detail} are based on two cycles being included in the input clip. We show that CSS can be used regardless of the number of cycles present and extend our examples to three cycles in Fig. \ref{detail_mult}, showing that the same CSS loss function can still be used. 

If there are three cardiac cycles in the input clip and two matching phases in the search region, we can still calculate matching probability at different time-points in $Q$ based on Equation \ref{eq:search_match}. Given that the features are jointly trained with supervised segmentation, the matching probabilities should be higher at the true matching time-points compared with other regions for well trained features. We use the same probability weightings to calculate expected value $\tilde{E}_i^{q^*+c}$ as per Equation~\ref{eq:expeted_match}. We illustrate this in Fig. \ref{detail_b_mult}. 

We can also perform phase matching back in the template region by using Equation \ref{eq:template_match} regardless of how many cycles are present in the input clip. This is because $\tilde{E}_i^{q^*+c}$ should still match with the phase at $p^*+c$ even if multiple cycles are used to calculate the expectation. Thus, we can use the same CSS loss formulation even for multiple cycles. 

One important detail to note is that having multiple phases \textit{in the template region} $P$ can lead to ambiguous matching for Equation \ref{eq:template_match}. Thus, we use a shorter interval for $P$ compared to $Q$ (15 frames \vs 21) to avoid such cases during implementation. This does not affect other components of our method since we only need $P$ for template sampling and matching, which does not require a full cycle. }

\begin{table}[h!]

    \captionsetup{labelfont={color=CLRBlue}}
  \caption{\wdairv{LVEF prediction results on EchoNet-Dynamic using different backbone architectures. We use one-eighth of available labels as the labeled dataset and treat remaining samples as unlabeled data. We report mean results $\pm$ standard deviation of five separate runs. 
  }}
  \label{tab:backbone}
  \centering
  
    \addtolength{\tabcolsep}{-3pt}    
 
  \color{CLRBlue}
    \begin{tabular}{c|c|c|cc}
    \toprule[1.5pt] 
    

    \multirow{1}{*}{Type} & \multirow{1}{*}{Method} & \multirow{1}{*}{Backbone} & MAE $\downarrow$ & $\mathbf{R}^2$ Values $\uparrow$ \\

\hline

\multirow{2}{*}{Supervised} & Ouyang \etal~\cite{ouyang2020video}      & MC3  & 6.04 \wdairv{$\pm$ 0.06} &  58.6\%  \wdairv{$\pm$ 0.5}   \\
 & Dai \etal~\cite{dai2021adaptive} & MC3   &  5.97    \wdairv{$\pm$ 0.05} &  60.1\%  \wdairv{$\pm$ 0.5}  \\

\hline

\multirow{3}{*}{\begin{tabular}[c]{@{}c@{}}Semi-\\Supervised\end{tabular}} & 
Ji \etal~\cite{ji2019learning}       & MC3  &  5.96 \wdairv{$\pm$ 0.05} & 60.4\% \wdairv{$\pm$ 0.5}  \\

 & Xu \etal~\cite{xu2021cross}      & MC3 &  5.95 \wdairv{$\pm$ 0.04} &  60.6\% \wdairv{$\pm$ 0.4}   \\

& Ours    & MC3 &\textbf{5.85 \wdairv{$\pm$ 0.04}} & \textbf{61.6\% \wdairv{$\pm$ 0.4}}   \\
\hline
\hline



\multirow{2}{*}{Supervised} & Ouyang \etal~\cite{ouyang2020video}      & R3D   & 5.90  \wdairv{$\pm$ 0.05}   & 60.0\% \wdairv{$\pm$ 0.5}  \\
 & Dai \etal~\cite{dai2021adaptive} & R3D    & 5.84 \wdairv{$\pm$ 0.04}  & 61.1\% \wdairv{$\pm$ 0.5} \\

\hline

\multirow{3}{*}{\begin{tabular}[c]{@{}c@{}}Semi-\\Supervised\end{tabular}} & 
Ji \etal~\cite{ji2019learning}       & R3D  & 5.75 \wdairv{$\pm$ 0.05} & 61.4\% \wdairv{$\pm$ 0.5}\\

 & Xu \etal~\cite{xu2021cross}      & R3D   & 5.69 \wdairv{$\pm$ 0.04} & 63.5\% \wdairv{$\pm$ 0.4} \\

& Ours    & R3D  & \textbf{5.50 \wdairv{$\pm$ 0.04}} & \textbf{66.5\% \wdairv{$\pm$ 0.4}}\\

\hline
\hline




\multirow{2}{*}{Supervised} & Ouyang \etal~\cite{ouyang2020video}      & R2+1D    & 5.64 \wdairv{$\pm$ 0.08} & 60.6\%  \wdairv{$\pm$ 0.8}   \\
 & Dai \etal~\cite{dai2021adaptive} & R2+1D    &  5.47    \wdairv{$\pm$ 0.07}  & 62.8\% \wdairv{$\pm$ 0.6}   \\

\hline

\multirow{3}{*}{\begin{tabular}[c]{@{}c@{}}Semi-\\Supervised\end{tabular}} & 
Ji \etal~\cite{ji2019learning}       & R2+1D &  5.61 \wdairv{$\pm$ 0.07}  & 61.6\% \wdairv{$\pm$ 0.6} \\

 & Xu \etal~\cite{xu2021cross}      & R2+1D   &  5.08 \wdairv{$\pm$ 0.04} & 66.8\% \wdairv{$\pm$ 0.4}  \\

& Ours    & R2+1D  &\textbf{4.90 \wdairv{$\pm$ 0.04}}  & \textbf{71.1\% \wdairv{$\pm$ 0.4}} \\

\bottomrule[1.5pt]
    
  \end{tabular}
    \addtolength{\tabcolsep}{3pt}    
    
\end{table}

\wdairv{

\section{Choice of input length for $X_i$} \label{apdx_cliplength}

For $X_i$, we use an input clip 40 frames in length, sampled at a rate of 1 in every 3 frames. We show that this length covers at least two cardiac cycles for most sequences in EchoNet-Dynamic. 

We note that the human pulse rate is typically between 50 beats-per-minute (bpm) and 90 bpm \cite{nanchen2018resting}, and that 79\% of all video sequences in EchoNet-Dynamic are captured at 50 frames-per-second (fps) \cite{ouyang2019echonet}. Therefore, the upper bound of frames required for two full cycles is approximately:
\begin{equation}
    \frac{60 \textrm{ seconds}}{50 \textrm{bpm}} \times 2 \textrm{ beats} \times 50 \textrm{ fps} = 120 \textrm{ frames}\:.
\end{equation}
Sampling 40 frames at a rate of 1 in every 3 frames covers $40 \times 3 - 2 = 118$ frames from the original sequence, which is most of the two full cardiac cycles. Furthermore, using 40 frames as the input clip length means that the CSS and supervised segmentation task are balanced in the number of frames used. This setting also allows for easier comparison with other semi-supervised methods that use 40 unlabeled frames per iteration. 

}

\wdairv{
\section{Choice of Backbone Architecture} \label{apdx_backbone}

Ouyang \etal compared 3D (R3D), Mixed Convolution (MC3), and R2+1D \cite{tran2018closer} ResNet architectures for LVEF regression and found that the R2+1D ResNet performed the best \cite{ouyang2019echonet,ouyang2020video}. The R2+1D backbone was also used as the baseline architecture in works by Dai \etal{} \cite{dai2021adaptive}. We repeat the experiments using one-eighth of available labels in Section \ref{sec:sota} and compare results for different backbone architectures in Table \ref{tab:backbone}. We can see the R2+1D backbone gives the best performance for each method, and therefore use this architecture for experiments. We also see our proposed method (Ours) outperforms alternative state-of-the-art approaches regardless of the backbone used.}

\begin{table}[h!]
  \caption{Summary statistics for EchoNet-Dynamic dataset.}
  \label{lvef_stat}
  \centering
  {
  \begin{tabular}{c|c}
    \toprule[1.5pt]
    Metric         & Value (Standard Deviation)\\
    \hline
    Total number of patients     & 10,030  \\
    Average number of frames & 175 (57) \\
    \wdairv{Average frames per second} & \wdairv{51.1 (6.2)} \\
    Mean LVEF \%     & 55.75 (12.37)   \\
    \bottomrule[1.5pt]
  \end{tabular} 
  }
\end{table}

\begin{table}[h!]

  \caption{Summary statistics for CAMUS dataset.}
  \label{lvef_camus}
  \centering
  { 
  \begin{tabular}{c|c}
    \toprule[1.5pt]
    Metric        & Value (Standard Deviation)\\
    \hline
    Total number of patients     & 450  \\
    Average number of frames & 20 (4) \\
    Video quality & Good: 260, Medium: 148, Poor: 42 \\
    Mean LVEF \%     & 52.03 (12.07)   \\
    \bottomrule[1.5pt]
  \end{tabular} 
  }
\end{table}

\wdairvb{
\section{Temporal Mirroring for CAMUS sequences} \label{apdx_mirror}

Temporal mirroring is done by reversing the sequences at the starting and ending time-points. 
For video video sequence $V$ consisting for frames $[v_1, v_2, ... , v_{T-1}, v_{T}]$,
the doubled sequence $V'$ is equal to:
\begin{equation}
\begin{split}
    V' & = [v_{T/2}, v_{T/2-1}, ..., v_2, v_1, v_2, ... , \\
    & v_{T-1}, v_{T}, v_{T-1}, ..., v_{T/2+1},v_{T/2}] \:.
\end{split}
\end{equation}
}

\begin{figure}%
\captionsetup{labelfont={color=CLRBlue},font={color=CLRBlue}}
\centering
\includegraphics[width=0.95 \columnwidth]{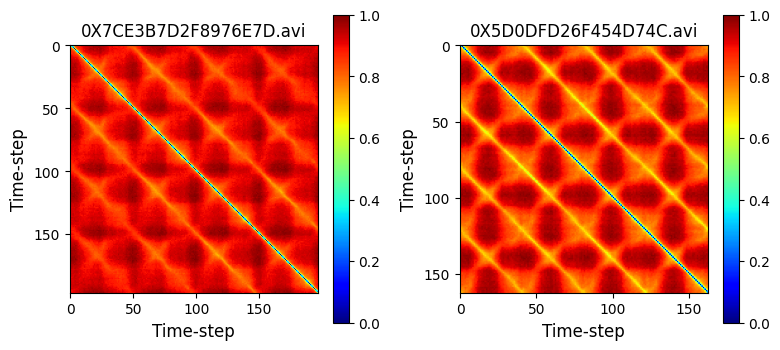}%

\caption{ Heatmap plot of normalized L2 distance between frames at different time-steps. 
Diagonal patterns of low L2 distance indicate presence of cyclical repetition since similar frames are repeated at regular intervals. 
}
\label{frame_sim}
\end{figure}

\begin{figure*}%
\captionsetup{labelfont={color=CLRBlue},font={color=CLRBlue}}
\centering
\includegraphics[width=2 \columnwidth]{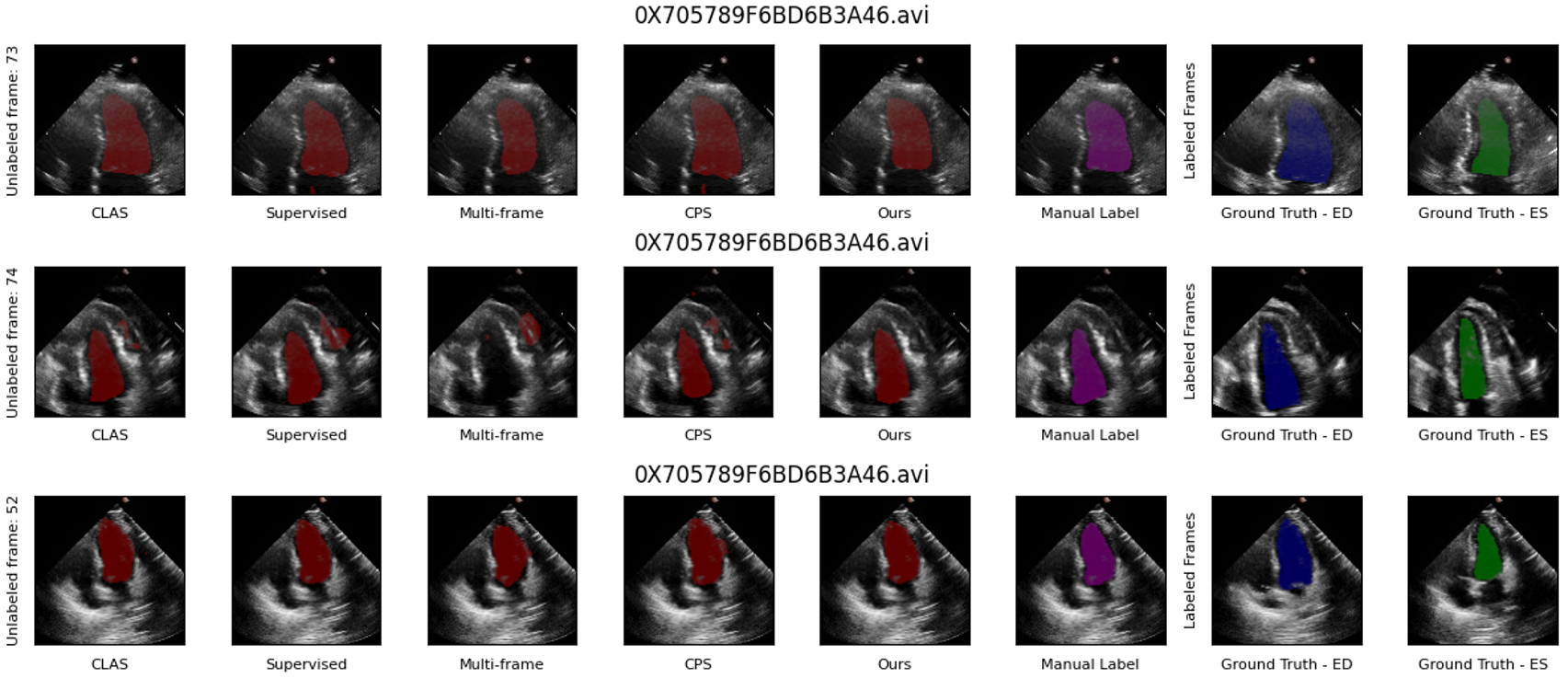}%

\caption{ Qualitative samples of LV mask predictions (\textcolor{red}{red}) on unlabeled non-ED and non-ES frames for different methods. Performance is evaluated by comparing with manual LV segmentations (\textcolor{Plum}{purple}). We show ground truth labels on ED (\textcolor{blue}{blue}) and ES (\textcolor{teal}{green}) frames for reference. We note that alternative methods tend to have predictions distinctly outside of the true LV region for difficult samples.  
}
\label{unlb_seg_examp}
\end{figure*}

\newpage

\bibliographystyle{IEEEtranN}
\small{\bibliography{refs}}
\end{document}

%% file: math_commands.tex
\usepackage{amsmath,amsfonts,bm}

\def\eqref#1{equation~\ref{#1}}

\def\1{\bm{1}}

\def\vs{{\bm{s}}}

\DeclareMathAlphabet{\mathsfit}{\encodingdefault}{\sfdefault}{m}{sl}
\SetMathAlphabet{\mathsfit}{bold}{\encodingdefault}{\sfdefault}{bx}{n}














%% file: tmi.bbl
\begin{thebibliography}{60}
\providecommand{\natexlab}[1]{#1}
\providecommand{\url}[1]{#1}
\csname url@samestyle\endcsname
\providecommand{\newblock}{\relax}
\providecommand{\bibinfo}[2]{#2}
\providecommand{\BIBentrySTDinterwordspacing}{\spaceskip=0pt\relax}
\providecommand{\BIBentryALTinterwordstretchfactor}{4}
\providecommand{\BIBentryALTinterwordspacing}{\spaceskip=\fontdimen2\font plus
\BIBentryALTinterwordstretchfactor\fontdimen3\font minus
  \fontdimen4\font\relax}
\providecommand{\BIBforeignlanguage}[2]{{%
\expandafter\ifx\csname l@#1\endcsname\relax
\typeout{** WARNING: IEEEtranN.bst: No hyphenation pattern has been}%
\typeout{** loaded for the language `#1'. Using the pattern for}%
\typeout{** the default language instead.}%
\else
\language=\csname l@#1\endcsname
\fi
#2}}
\providecommand{\BIBdecl}{\relax}
\BIBdecl

\bibitem[Bamira and Picard(2018)]{bamira2018imaging}
D.~Bamira and M.~Picard, ``Imaging: Echocardiology—assessment of cardiac
  structure and function,'' \emph{Encyclopedia of Cardiovascular Research and
  Medicine}, pp. 35--54, 2018.

\bibitem[Maeder and Kaye(2009)]{maeder2009heart}
M.~T. Maeder and D.~M. Kaye, ``Heart failure with normal left ventricular
  ejection fraction,'' \emph{Journal of the American College of Cardiology},
  vol.~53, no.~11, pp. 905--918, 2009.

\bibitem[Abdi~\etal(2017)]{abdi2017automatic}
A.~H. Abdi~\etal, ``Automatic quality assessment of echocardiograms using
  convolutional neural networks: feasibility on the apical four-chamber view,''
  \emph{IEEE transactions on medical imaging}, vol.~36, no.~6, pp. 1221--1230,
  2017.

\bibitem[Hughes~\etal(2021)]{hughes2021deep}
J.~W. Hughes~\etal, ``Deep learning evaluation of biomarkers from
  echocardiogram videos,'' \emph{EBioMedicine}, vol.~73, p. 103613, 2021.

\bibitem[Ouyang~\etal(2019)]{ouyang2019echonet}
D.~Ouyang~\etal, ``Echonet-dynamic: a large new cardiac motion video data
  resource for medical machine learning,'' in \emph{NeurIPS}, 2019.

\bibitem[Ouyang~\textit{et al.}(2020)]{ouyang2020video}
D.~Ouyang~\textit{et al.}, ``Video-based ai for beat-to-beat assessment of
  cardiac function,'' \emph{Nature}, vol. 580, no. 7802, pp. 252--256, 2020.

\bibitem[Cole~\etal(2015)]{cole2015defining}
G.~D. Cole~\etal, ``Defining the real-world reproducibility of visual grading
  of left ventricular function and visual estimation of left ventricular
  ejection fraction: impact of image quality, experience and accreditation,''
  \emph{The international journal of cardiovascular imaging}, vol.~31, no.~7,
  pp. 1303--1314, 2015.

\bibitem[Jafari et~al.(2021)Jafari, Van~Woudenberg, Luong, Abolmaesumi, and
  Tsang]{jafari2021deep}
M.~H. Jafari, N.~Van~Woudenberg, C.~Luong, P.~Abolmaesumi, and T.~Tsang, ``Deep
  bayesian image segmentation for a more robust ejection fraction estimation,''
  in \emph{ISBI}.\hskip 1em plus 0.5em minus 0.4em\relax IEEE, 2021, pp.
  1264--1268.

\bibitem[Reynaud et~al.(2021)Reynaud, Vlontzos, Hou, Beqiri, Leeson, and
  Kainz]{reynaud2021ultrasound}
H.~Reynaud, A.~Vlontzos, B.~Hou, A.~Beqiri, P.~Leeson, and B.~Kainz,
  ``Ultrasound video transformers for cardiac ejection fraction estimation,''
  \emph{arXiv preprint arXiv:2107.00977}, 2021.

\bibitem[Wei et~al.(2020)Wei, Cao, Cao, Zhou, Xue, Ni, and Li]{wei2020temporal}
H.~Wei, H.~Cao, Y.~Cao, Y.~Zhou, W.~Xue, D.~Ni, and S.~Li,
  ``Temporal-consistent segmentation of echocardiography with co-learning from
  appearance and shape,'' in \emph{Medical Image Computing and Computer
  Assisted Intervention--MICCAI 2020: 23rd International Conference, Lima,
  Peru, October 4--8, 2020, Proceedings, Part II}, 2020, pp. 623--632.

\bibitem[Yuan~\etal(2021)]{yuan2021systematic}
N.~Yuan~\etal, ``Systematic quantification of sources of variation in ejection
  fraction calculation using deep learning,'' \emph{Cardiovascular Imaging},
  vol.~14, no.~11, pp. 2260--2262, 2021.

\bibitem[Dai et~al.(2021)Dai, Li, Chiu, Kuo, and Cheng]{dai2021adaptive}
W.~Dai, X.~Li, W.~H.~K. Chiu, M.~D. Kuo, and K.-T. Cheng, ``Adaptive contrast
  for image regression in computer-aided disease assessment,'' \emph{IEEE
  Transactions on Medical Imaging}, 2021.

\bibitem[Ghorbani~\etal(2020)]{ghorbani2020deep}
A.~Ghorbani~\etal, ``Deep learning interpretation of echocardiograms,''
  \emph{NPJ digital medicine}, vol.~3, no.~1, pp. 1--10, 2020.

\bibitem[Leclerc~\textit{et al.}(2019)]{leclerc2019deep}
S.~Leclerc~\textit{et al.}, ``Deep learning for segmentation using an open
  large-scale dataset in 2d echocardiography,'' \emph{IEEE transactions on
  medical imaging}, vol.~38, no.~9, pp. 2198--2210, 2019.

\bibitem[Liu et~al.(2021{\natexlab{a}})Liu, Wang, Liu, Yang, and
  Tian]{liu2021deep}
F.~Liu, K.~Wang, D.~Liu, X.~Yang, and J.~Tian, ``Deep pyramid local attention
  neural network for cardiac structure segmentation in two-dimensional
  echocardiography,'' \emph{Medical Image Analysis}, vol.~67, p. 101873, 2021.

\bibitem[Painchaud et~al.(2021)Painchaud, Duchateau, Bernard, and
  Jodoin]{painchaud2021echocardiography}
N.~Painchaud, N.~Duchateau, O.~Bernard, and P.-M. Jodoin, ``Echocardiography
  segmentation with enforced temporal consistency,'' \emph{arXiv preprint
  arXiv:2112.02102}, 2021.

\bibitem[Chen et~al.(2022)Chen, Zhang, Haggerty, and Stough]{chen2022fully}
Y.~Chen, X.~Zhang, C.~M. Haggerty, and J.~V. Stough, ``Fully automated
  multi-heartbeat echocardiography video segmentation and motion tracking,'' in
  \emph{Medical Imaging 2022: Image Processing}, vol. 12032.\hskip 1em plus
  0.5em minus 0.4em\relax SPIE, 2022, pp. 185--192.

\bibitem[Dezaki~\etal(2018)]{dezaki2018cardiac}
F.~T. Dezaki~\etal, ``Cardiac phase detection in echocardiograms with densely
  gated recurrent neural networks and global extrema loss,'' \emph{IEEE
  transactions on medical imaging}, vol.~38, no.~8, pp. 1821--1832, 2018.

\bibitem[Tran et~al.(2018)Tran, Wang, Torresani, Ray, LeCun, and
  Paluri]{tran2018closer}
D.~Tran, H.~Wang, L.~Torresani, J.~Ray, Y.~LeCun, and M.~Paluri, ``A closer
  look at spatiotemporal convolutions for action recognition,'' in \emph{CVPR},
  2018, pp. 6450--6459.

\bibitem[Tarvainen and Valpola(2017)]{tarvainen2017mean}
A.~Tarvainen and H.~Valpola, ``Mean teachers are better role models:
  Weight-averaged consistency targets improve semi-supervised deep learning
  results,'' \emph{arXiv preprint arXiv:1703.01780}, 2017.

\bibitem[Li et~al.(2018)Li, Yu, Chen, Fu, and Heng]{li2018semi}
X.~Li, L.~Yu, H.~Chen, C.-W. Fu, and P.-A. Heng, ``Semi-supervised skin lesion
  segmentation via transformation consistent self-ensembling model,'' in
  \emph{BMVC}, 2018.

\bibitem[Yu et~al.(2019)Yu, Wang, Li, Fu, and Heng]{yu2019uncertainty}
L.~Yu, S.~Wang, X.~Li, C.-W. Fu, and P.-A. Heng, ``Uncertainty-aware
  self-ensembling model for semi-supervised 3d left atrium segmentation,'' in
  \emph{MICCAI}.\hskip 1em plus 0.5em minus 0.4em\relax Springer, 2019, pp.
  605--613.

\bibitem[Ouali et~al.(2020)Ouali, Hudelot, and Tami]{ouali2020semi}
Y.~Ouali, C.~Hudelot, and M.~Tami, ``Semi-supervised semantic segmentation with
  cross-consistency training,'' in \emph{CVPR}, 2020, pp. 12\,674--12\,684.

\bibitem[Li et~al.(2020)Li, Yu, Chen, Fu, Xing, and Heng]{li2020transformation}
X.~Li, L.~Yu, H.~Chen, C.-W. Fu, L.~Xing, and P.-A. Heng,
  ``Transformation-consistent self-ensembling model for semisupervised medical
  image segmentation,'' \emph{IEEE Transactions on Neural Networks and Learning
  Systems}, vol.~32, no.~2, pp. 523--534, 2020.

\bibitem[Chen et~al.(2021)Chen, Yuan, Zeng, and Wang]{chen2021semi}
X.~Chen, Y.~Yuan, G.~Zeng, and J.~Wang, ``Semi-supervised semantic segmentation
  with cross pseudo supervision,'' in \emph{CVPR}, 2021, pp. 2613--2622.

\bibitem[Yao et~al.(2022)Yao, Hu, and Li]{yao2022enhancing}
H.~Yao, X.~Hu, and X.~Li, ``Enhancing pseudo label quality for semi-supervised
  domain-generalized medical image segmentation,'' in \emph{AAAI}, 2022.

\bibitem[Lin et~al.(2022)Lin, Yao, Li, Zheng, and Li]{lin2022calibrating}
Y.~Lin, H.~Yao, Z.~Li, G.~Zheng, and X.~Li, ``Calibrating label distribution
  for class-imbalanced barely-supervised knee segmentation,'' in \emph{MICCAI},
  2022.

\bibitem[Wu et~al.(2021)Wu, Li, and Cheng]{wu2021exploring}
H.~Wu, X.~Li, and K.-T. Cheng, ``Exploring feature representation learning for
  semi-supervised medical image segmentation,'' \emph{arXiv preprint
  arXiv:2111.10989}, 2021.

\bibitem[Alonso et~al.(2021)Alonso, Sabater, Ferstl, Montesano, and
  Murillo]{alonso2021semi}
I.~Alonso, A.~Sabater, D.~Ferstl, L.~Montesano, and A.~C. Murillo,
  ``Semi-supervised semantic segmentation with pixel-level contrastive learning
  from a class-wise memory bank,'' \emph{arXiv preprint arXiv:2104.13415},
  2021.

\bibitem[Liu et~al.(2021{\natexlab{b}})Liu, Zhi, Johns, and
  Davison]{liu2021bootstrapping}
S.~Liu, S.~Zhi, E.~Johns, and A.~J. Davison, ``Bootstrapping semantic
  segmentation with regional contrast,'' \emph{arXiv preprint
  arXiv:2104.04465}, 2021.

\bibitem[Zou~\etal(2020)]{zou2020pseudoseg}
Y.~Zou~\etal, ``Pseudoseg: Designing pseudo labels for semantic segmentation,''
  \emph{arXiv preprint arXiv:2010.09713}, 2020.

\bibitem[Yan~\etal(2019)]{yan2019semi}
P.~Yan~\etal, ``Semi-supervised video salient object detection using
  pseudo-labels,'' in \emph{CVPR}, 2019, pp. 7284--7293.

\bibitem[Varghese~\etal(2020)]{varghese2020unsupervised}
S.~Varghese~\etal, ``Unsupervised temporal consistency metric for video
  segmentation in highly-automated driving,'' in \emph{CVPR}, 2020, pp.
  336--337.

\bibitem[Ding~\etal(2020)]{ding2020every}
M.~Ding~\etal, ``Every frame counts: joint learning of video segmentation and
  optical flow,'' in \emph{AAAI}, vol.~34, no.~07, 2020, pp. 10\,713--10\,720.

\bibitem[Perazzi et~al.(2017)Perazzi, Khoreva, Benenson, Schiele, and
  Sorkine-Hornung]{perazzi2017learning}
F.~Perazzi, A.~Khoreva, R.~Benenson, B.~Schiele, and A.~Sorkine-Hornung,
  ``Learning video object segmentation from static images,'' in \emph{CVPR},
  2017, pp. 2663--2672.

\bibitem[Oh et~al.(2019)Oh, Lee, Xu, and Kim]{oh2019video}
S.~W. Oh, J.-Y. Lee, N.~Xu, and S.~J. Kim, ``Video object segmentation using
  space-time memory networks,'' in \emph{CVPR}, 2019, pp. 9226--9235.

\bibitem[Cheng et~al.(2021)Cheng, Tai, and Tang]{cheng2021rethinking}
H.~K. Cheng, Y.-W. Tai, and C.-K. Tang, ``Rethinking space-time networks with
  improved memory coverage for efficient video object segmentation,''
  \emph{arXiv preprint arXiv:2106.05210}, 2021.

\bibitem[Zhang~\etal(2020)]{zhang2020semi}
Y.~Zhang~\etal, ``Semi-supervised cardiac image segmentation via label
  propagation and style transfer,'' in \emph{STACOM}.\hskip 1em plus 0.5em
  minus 0.4em\relax Springer, 2020, pp. 219--227.

\bibitem[Pedrosa~\textit{et al.}(2017)]{pedrosa2017fast}
J.~Pedrosa~\textit{et al.}, ``Fast and fully automatic left ventricular
  segmentation and tracking in echocardiography using shape-based b-spline
  explicit active surfaces,'' \emph{IEEE transactions on medical imaging},
  vol.~36, no.~11, pp. 2287--2296, 2017.

\bibitem[Ta et~al.(2020)Ta, Ahn, Stendahl, Sinusas, and Duncan]{ta2020semi}
K.~Ta, S.~S. Ahn, J.~C. Stendahl, A.~J. Sinusas, and J.~S. Duncan, ``A
  semi-supervised joint network for simultaneous left ventricular motion
  tracking and segmentation in 4d echocardiography,'' in \emph{MICCAI}.\hskip
  1em plus 0.5em minus 0.4em\relax Springer, 2020, pp. 468--477.

\bibitem[Qin~\textit{et al.}(2018)]{qin2018joint}
C.~Qin~\textit{et al.}, ``Joint learning of motion estimation and segmentation
  for cardiac mr image sequences,'' in \emph{MICCAI}.\hskip 1em plus 0.5em
  minus 0.4em\relax Springer, 2018, pp. 472--480.

\bibitem[Xiong et~al.(2021)Xiong, Fan, Grauman, and
  Feichtenhofer]{xiong2021multiview}
B.~Xiong, H.~Fan, K.~Grauman, and C.~Feichtenhofer, ``Multiview pseudo-labeling
  for semi-supervised learning from video,'' \emph{arXiv preprint
  arXiv:2104.00682}, 2021.

\bibitem[Jing et~al.(2021)Jing, Parag, Wu, Tian, and Wang]{jing2021videossl}
L.~Jing, T.~Parag, Z.~Wu, Y.~Tian, and H.~Wang, ``Videossl: Semi-supervised
  learning for video classification,'' in \emph{WACV}, 2021, pp. 1110--1119.

\bibitem[Xu~\textit{et al.}(2021)]{xu2021cross}
Y.~Xu~\textit{et al.}, ``Cross-model pseudo-labeling for semi-supervised action
  recognition,'' \emph{arXiv preprint arXiv:2112.09690}, 2021.

\bibitem[Ding et~al.(2021)Ding, Wang, Gao, Li, Wang, and Liu]{ding2021kfc}
X.~Ding, N.~Wang, X.~Gao, J.~Li, X.~Wang, and T.~Liu, ``Kfc: An efficient
  framework for semi-supervised temporal action localization,'' \emph{IEEE
  Transactions on Image Processing}, vol.~30, pp. 6869--6878, 2021.

\bibitem[Ji et~al.(2019)Ji, Cao, and Niebles]{ji2019learning}
J.~Ji, K.~Cao, and J.~C. Niebles, ``Learning temporal action proposals with
  fewer labels,'' in \emph{CVPR}, 2019, pp. 7073--7082.

\bibitem[Hu et~al.(2021)Hu, Zeng, Xu, and Shi]{hu2021semi}
X.~Hu, D.~Zeng, X.~Xu, and Y.~Shi, ``Semi-supervised contrastive learning for
  label-efficient medical image segmentation,'' in \emph{International
  Conference on Medical Image Computing and Computer-Assisted
  Intervention}.\hskip 1em plus 0.5em minus 0.4em\relax Springer, 2021, pp.
  481--490.

\bibitem[Ouyang et~al.(2015)Ouyang, Li, Zeng, and Wang]{ouyang2015learning}
W.~Ouyang, H.~Li, X.~Zeng, and X.~Wang, ``Learning deep representation with
  large-scale attributes,'' in \emph{Proceedings of the IEEE International
  Conference on Computer Vision}, 2015, pp. 1895--1903.

\bibitem[Chen et~al.(2017)Chen, Papandreou, Schroff, and
  Adam]{chen2017rethinking}
L.-C. Chen, G.~Papandreou, F.~Schroff, and H.~Adam, ``Rethinking atrous
  convolution for semantic image segmentation,'' \emph{arXiv preprint
  arXiv:1706.05587}, 2017.

\bibitem[He et~al.(2016)He, Zhang, Ren, and Sun]{he2016deep}
K.~He, X.~Zhang, S.~Ren, and J.~Sun, ``Deep residual learning for image
  recognition,'' in \emph{CVPR}, 2016, pp. 770--778.

\bibitem[Kay~\textit{et al.}(2017)]{kay2017kinetics}
W.~Kay~\textit{et al.}, ``The kinetics human action video dataset,''
  \emph{arXiv preprint arXiv:1705.06950}, 2017.

\bibitem[Smilkov et~al.(2017)Smilkov, Thorat, Kim, Vi{\'e}gas, and
  Wattenberg]{smilkov2017smoothgrad}
D.~Smilkov, N.~Thorat, B.~Kim, F.~Vi{\'e}gas, and M.~Wattenberg, ``Smoothgrad:
  removing noise by adding noise,'' \emph{arXiv preprint arXiv:1706.03825},
  2017.

\bibitem[Dwibedi et~al.(2019)Dwibedi, Aytar, Tompson, Sermanet, and
  Zisserman]{dwibedi2019temporal}
D.~Dwibedi, Y.~Aytar, J.~Tompson, P.~Sermanet, and A.~Zisserman, ``Temporal
  cycle-consistency learning,'' in \emph{Proceedings of the IEEE/CVF conference
  on computer vision and pattern recognition}, 2019, pp. 1801--1810.

\bibitem[Dezaki~\textit{et al.}(2021)]{dezaki2021echo}
F.~T. Dezaki~\textit{et al.}, ``Echo-syncnet: Self-supervised cardiac view
  synchronization in echocardiography,'' \emph{IEEE Transactions on Medical
  Imaging}, vol.~40, no.~8, pp. 2092--2104, 2021.

\bibitem[Ahn et~al.(2021)Ahn, Ta, Thorn, Langdon, Sinusas, and
  Duncan]{ahn2021multi}
S.~S. Ahn, K.~Ta, S.~Thorn, J.~Langdon, A.~J. Sinusas, and J.~S. Duncan,
  ``Multi-frame attention network for left ventricle segmentation in 3d
  echocardiography,'' in \emph{MICCAI}.\hskip 1em plus 0.5em minus 0.4em\relax
  Springer, 2021, pp. 348--357.

\bibitem[Leeson et~al.(2012)Leeson, Augustine, Mitchell, and
  Becher]{leeson2012echocardiography}
P.~Leeson, D.~Augustine, A.~R. Mitchell, and H.~Becher,
  \emph{Echocardiography}.\hskip 1em plus 0.5em minus 0.4em\relax Oxford
  University Press, 2012.

\bibitem[Zhong et~al.(2022)Zhong, Tang, Chen, Peng, and Wang]{zhong2022self}
Y.~Zhong, H.~Tang, J.~Chen, J.~Peng, and Y.-X. Wang, ``Is self-supervised
  learning more robust than supervised learning?'' \emph{arXiv preprint
  arXiv:2206.05259}, 2022.

\bibitem[Hendrycks et~al.(2019)Hendrycks, Mazeika, Kadavath, and
  Song]{hendrycks2019using}
D.~Hendrycks, M.~Mazeika, S.~Kadavath, and D.~Song, ``Using self-supervised
  learning can improve model robustness and uncertainty,'' \emph{Advances in
  neural information processing systems}, vol.~32, 2019.

\bibitem[Tse(2016)]{tse2016mechanisms}
G.~Tse, ``Mechanisms of cardiac arrhythmias,'' \emph{Journal of arrhythmia},
  vol.~32, no.~2, pp. 75--81, 2016.

\bibitem[Nanchen(2018)]{nanchen2018resting}
D.~Nanchen, ``Resting heart rate: what is normal?'' pp. 1048--1049, 2018.

\end{thebibliography}
